\documentclass[journal,twoside]{IEEEtran}
\usepackage{multirow}
\usepackage{multicol}
\usepackage{amssymb}
\usepackage{makecell}
\usepackage{hhline}
\usepackage{colortbl}
\definecolor{Gray}{gray}{0.85}
\usepackage{ifpdf}
\usepackage{cite}
\usepackage{hyperref}
\hypersetup{
    colorlinks=true,
    citecolor=black,
    linkcolor=black,
    filecolor=black,      
    urlcolor=black,
}

\ifCLASSINFOpdf
   \usepackage[pdftex]{graphicx}
   \graphicspath{{../pdf/}{../jpeg/}}
   \DeclareGraphicsExtensions{.pdf,.jpeg,.png}
\else
   \usepackage[dvips]{graphicx}
   \graphicspath{{../eps/}}
   \DeclareGraphicsExtensions{.eps}
\fi
\usepackage{amsmath}
\usepackage{algorithmic}
\usepackage{array}
\ifCLASSOPTIONcompsoc
  \usepackage[caption=false,font=normalsize,labelfont=sf,textfont=sf]{subfig}
\else
 \usepackage[caption=false,font=footnotesize]{subfig}
\fi
\usepackage{fixltx2e}
 \usepackage{dblfloatfix}
\ifCLASSOPTIONcaptionsoff
  \usepackage[nomarkers]{endfloat}
 \let\MYoriglatexcaption\caption
 \renewcommand{\caption}[2][\relax]{\MYoriglatexcaption[#2]{#2}}
\fi
\usepackage{url}
\begin{document}
\title{Online Multi-Object Tracking and Segmentation with GMPHD Filter\\ and Mask-based Affinity Fusion}

\author{Young-min~Song,~Young-chul~Yoon,~Kwangjin~Yoon,\\
Moongu~Jeon*,~Seong-Whan~Lee,~and~Witold~Pedrycz% <-this % stops a space
%\thanks{Manuscript received April 26, 2021; revised April 26, 2021. 
\thanks{Y. Song and M. Jeon are with the School of Electrical Engineering and Computer Science, Gwangju Institute of Technology,
Gwangju 61005, South Korea (e-mail: sym@gist.ac.kr; mgjeon@gist.ac.kr).}% <-this % stops a space
\thanks{Y. Yoon is with the Robotics Laboratory, Hyundai Motor Company, Gyeonggi-do 16082, South Korea (e-mail: youngchul.yoon@hyundai.com).}% <-this % stops a space
\thanks{K. Yoon is with SI Analytics Co., Ltd., Daejeon 34051, South Korea 
(e-mail: yoon28@si-analytics.ai).}% <-this % stops a space
\thanks{Seong-Whan Lee is with the Department of Artificial Intelligence, Korea University, Seoul 02841, South Korea, and also with the Department of Brain and Cognitive Engineering, Korea University, Seoul 02841, South Korea
(e-mail: sw.lee@korea.ac.kr).}% <-this % stops a space
\thanks{W. Pedrycz is with the Department of Electrical and Computer Engineering, University of Alberta, Edmonton, AB T6R 2H5, Canada, also with the Department of Electrical and Computer Engineering, Faculty of Engineering, King Abdulaziz University, Jeddah 21589, Saudi Arabia, and also with the Systems Research Institute, Polish Academy of Sciences, 01-447 Warsaw, Poland 
(e-mail: wpedrycz@ualberta.ca).}}% <-this % stops a space

% line to be shifted left slightly. This is one of those ``LaTeX things". For
% instance, ``\textbf{A} \textbf{B}" will typeset as ``A B" not ``AB". To get
% ``AB" then you have to do: ``\textbf{A}\textbf{B}"

%\markboth{Journal of \LaTeX\ Class Files,~Vol.~14, No.~8, May~2021}%
\markboth{}%
{Song \MakeLowercase{\textit{et al.}}: Online Multi-Object Tracking and Segmentation with GMPHD Filter and Mask based Affinity Fusion}



\maketitle

\begin{abstract}
    In this paper, we propose a highly practical fully online multi-object tracking and segmentation (MOTS) method that uses instance segmentation results as an input. The proposed method is based on the Gaussian mixture probability hypothesis density (GMPHD) filter, a hierarchical data association (HDA), and a mask-based affinity fusion (MAF) model to achieve high-performance online tracking. The HDA consists of two associations: segment-to-track and track-to-track associations.  
    One affinity, for position and motion, is computed by using the GMPHD filter, 
    and the other affinity, for appearance is computed by using the responses from a single object tracker such as a kernalized correlation filter. These two affinities are simply fused by using a score-level fusion method such as min-max normalization referred to as MAF. 
    In addition, to reduce the number of false positive segments, we adopt mask IoU-based merging (mask merging). The proposed MOTS framework with the key modules: HDA, MAF, and mask merging, is easily extensible to simultaneously track multiple types of objects with CPU-only execution in parallel processing. In addition, the developed framework only requires simple parameter tuning unlike many existing MOTS methods that need intensive hyperparameter optimization.
    In the experiments on the two popular MOTS datasets, the key modules show some improvements. 
    For instance, ID-switch decreases by more than half compared to a baseline method in the training sets. In conclusion, our tracker achieves state-of-the-art MOTS performance in the test sets.
\end{abstract}


% \fi
\IEEEpeerreviewmaketitle

\section{Introduction}
\label{sec:intro}
% Here we have the typical use of a ``T" for an initial drop letter
% and ``HIS" in caps to complete the first word.
\IEEEPARstart{M}{ulti-object} tracking and segmentation (MOTS) has recently become an active research field. Since the representative multi-object tracking (MOT) benchmark datasets~\cite{mot15,kitti,mot16} were released, the tracking-by-detection paradigm has been the mainstream for MOT. Additionally, breakthroughs in object detection have been achieved by many deep neural network (DNN)-based detectors~\cite{pointpillars,yolo,frcnn,pointrcnn} from various sensor domains, such as color cameras (2D images) and LiDAR (3D point clouds). 
According to the input source, the detectors give different outputs, i.e., observations.
For instance, the detection responses of~\cite{yolo,frcnn} are 2D bounding boxes and those of ~\cite{pointpillars,pointrcnn} are 3D boxes.
In addition, K. He~\textit{et al.}~\cite{maskrcnn} introduced a pixelwise classification and detection method represented by instance segmentation, which has motivated many segmentation-based studies.

\begin{figure}[t]
\begin{center}
   \includegraphics[width=8.5cm]{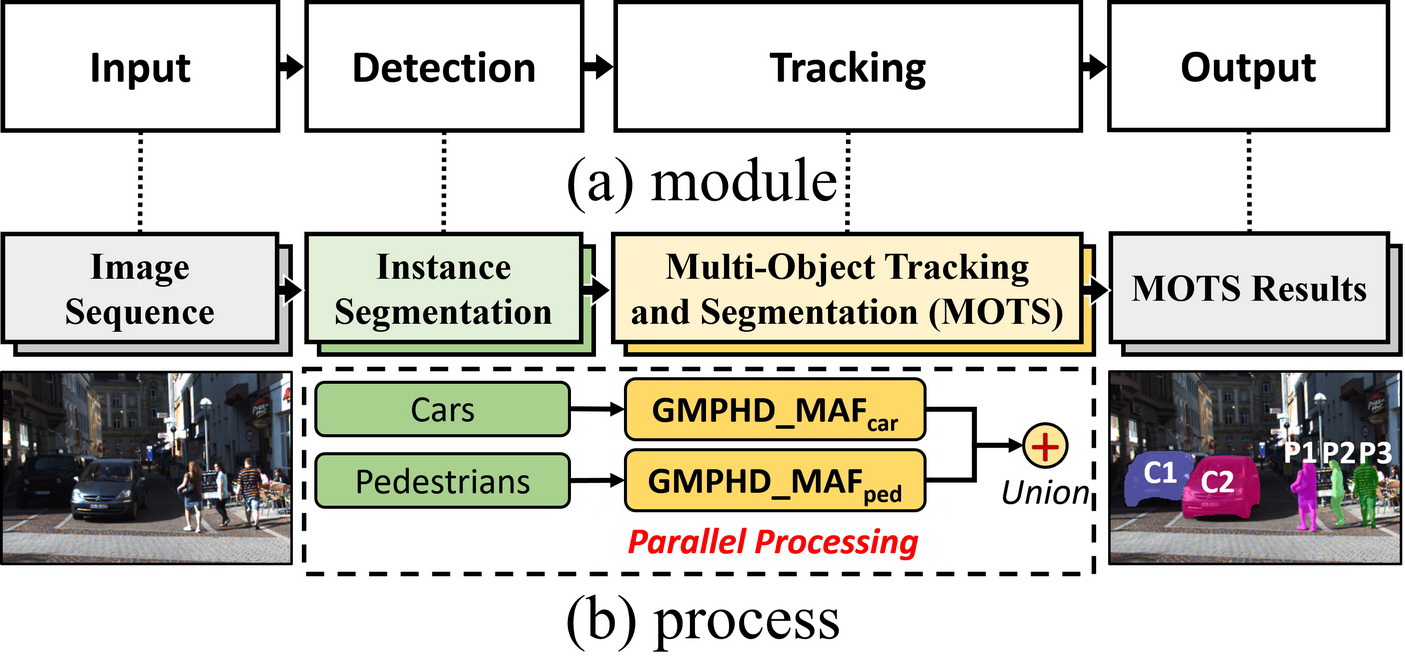}
\end{center}
   \caption{Flow chart of parallel multi-object tracking and segmentation (MOTS) processing, which receives two classes of objects, i.e., cars and pedestrians. (a) Modules (input, detection, tracking, and output) are implemented as (b) processes (image sequencing, instance segmentation, MOTS, and MOTS results). Our proposed MOTS framework is denoted by GMPHD\_MAF.}
\label{fig1}
\end{figure}

Recently, a new MOT method extended with segmentation, named MOTS has been developed for pixelwise intelligent systems beyond 2D bounding boxes and was first introduced in Voigtlaender~\textit{et al.}~\cite{mots_trackrcnn} with new evaluation measures and a new baseline method. They also released a new dataset extended from KITTI~\cite{kitti} and  MOTChallenge~\cite{mot16} image sequences. 
%기반을 다지고 새로운 연구주제를 만듬% 
Luiten~\textit{et al.}~\cite{MOTSFusionJ} proposed a MOTS method that uses a fusion of 2D box detection, 3D box detection, and instance segmentation results. Motivated by these MOTS works and other conventional MOT studies, in this paper, we propose a highly practical online MOTS method.

Our contributions are summarized as follows:
\begin{quotation}
    \noindent 1) We propose a highly practical online MOTS method consisting of (a) two-stage hierarchical data association (HDA), (b) mask-based affinity fusion (MAF), and (c) mask merging. These key modules can form the proposed online MOTS method with CPU-only execution.
    
    \noindent 2) Particularly among the key modules, (b) MAF effectively fuses ``position and motion" affinity with a Gaussian mixture probability density filter (GMPHD) and ``appearance" with a kernelized correlation filter (KCF) to improve the MOTS performance compared to a baseline method using only one-stage GMPHD filter association.
    
    \noindent 3) Additionally, the proposed method can be implemented with CPU-only execution so that it can run in parallel to simultaneously track multiple types of objects: cars and pedestrians, in this paper (see Figure~\ref{fig1}).
    % Also, the proposed method is feasible for the implementation in CPU-only so can run in parallel processing for simultaneously tracking multi-type objects; cars and pedestrians in this paper.
    
    \noindent 4) Finally, we evaluate the proposed method on state-of-the-art datasets~\cite{kitti,mot16,mots_trackrcnn}. The evaluation results on the training sets show incremental improvements compared to a baseline method. In the results on the test sets, our method not only shows competitive performance against state-of-the-art published methods but also achieves state-of-the-art level performance against state-of-the-art unpublished methods that are available at the leaderboards of the KITTI-MOTS and MOTSChallenge websites.
\end{quotation}

The proposed MOTS method has high applicability due to CPU-only execution and simple parameter tuning unlike many state-of-the-art DNN based tracking methods~\cite{mots_trackrcnn,camot,ciwt,beyondpixels,motsnet,MOTSFusionJ,PointTrack,ReMOTS} that need intensive hyperparameter optimization and heavy computing resources. In addition, in the experiments, our method shows state-of-the-art level performance.
We present the works related to the proposed method in Section~\ref{sec:related} and the details of the proposed one are covered in Section~\ref{sec:proposed}. Additionally, we discuss the experimental results in Section~\ref{sec:experiments} and conclude the paper in Section~\ref{sec:conclusion}. In what follows, we use GMPHD\_MAF as the abbreviation for the proposed method.

\section{Related Works}
\label{sec:related}
\subsection{Multi-Object Tracking with a PHD Filter}
% Filtering to 2D bb based MOT
% Vo,.. Bounding box 2D detection, my works.. Fu Duyong
The PHD filter~\cite{Mahler,Vo_gmphd,smcphd} was originally designed to deal with radar and sonar data-based MOT systems. Mahler~\textit{et al.}~\cite{Mahler} proposed recursive Bayes filter equations for the PHD filter that optimize multi-target tracking processes in radar and sonar systems with a random-finite set definition of states and observations. Following this PHD filtering theory, Vo~\textit{et al.}~\cite{smcphd} proposed a sequential Monte Carlo implementation of the PHD filter by using particle filtering and clustering, named the SMCPHD filter, and implemented the governing equations by using the Gaussian mixture model as a closed-form recursion method named the Gaussian mixture probability hypothesis density (GMPHD) filter. 

Since the GMPHD filter is tractable in implementing online and real-time trackers, it has been recently extended and exploited as a famous tracking model in video-based systems.
While the radar and sonar sensors receive massive number of false positives but rarely miss any observations, visual object detectors yield many fewer false positives and more missed detections. Thus, in video-based tracking, noise control processes for the original domains are simplified and many additional techniques for visual objects have been developed.
Song~\textit{et al.}~\cite{gmphdogm} extended GMPHD filter-based tracking with a two-stage hierarchical data association scheme to recover lost tracks IDs. They modeled an affinity function in the second stage association between the tracks by using the tracks' position and linear motion. This approach is an intuitive implementation of the GMPHD filter to reconnect lost tracks. In addition, they presented an energy minimization model based on occluded objects group to correct the false associations that already occurred in the first stage association between detections and tracks. 
Sanchez-Matilla~\textit{et al.}~\cite{eamtt} proposed detection confidence-based data association schemes with a PHD filter. Strong (high-confidence) detections initiate and propagate tracks, but weak (low-confidence) detections propagate only existing tracks. This scheme works well when the detection results are reliable. However, the tracking performance depends on the detection performance and is especially weak for long-term missed detections. 
More intensive solutions~\cite{gmphdkcf,fu1,phdgm} using appearance learning and motion modeling have been proposed. 
Kutschbach~\textit{et al.}~\cite{gmphdkcf} applied kernelized correlation filters (KCF)~\cite{kcf} to the naive GMPHD filtering process for online appearance updating to discriminate occluded objects. They proposed robust online appearance learning to refind the IDs of lost tracks. However, updating the appearance of every object in every frame requires heavy computing resources and inevitably increases the runtime. 
Fu~\textit{et al.}~\cite{fu1} added an adaptive gating technique and an online group-structured dictionary for appearance learning into the GMPHD filter. They made the GMPHD filter into a sophisticated tracking process suitable for video-based MOT. Sanchez-Matilla~\textit{et al.}~\cite{phdgm} proposed a global motion model based on long short-term memory (LSTM) models for MOT.
Some methods~\cite{gmphd2012,mtdf,gmphdn1tr} have proposed fusion methods to complement false positive and negative detections.
Kutschbach~\textit{et al.}~\cite{gmphd2012} developed a fusion model of a blob detector~\cite{blob} and head detector~\cite{head} in the GMPHD filter-based tracking. Fu~\textit{et al.}~\cite{mtdf} used a full-body detector~\cite{dpm} and body-part detector~\cite{pose} in their tracking-by-detection method. Baisa~\textit{et al.}~\cite{gmphdn1tr} proposed another type of fusion method that tracks multiple types of objects (cars and pedestrians) simultaneously by using an object recognition method such as a faster region-based convolutional neural network (FRCNN)~\cite{frcnn} in parallel processes.
These online MOT methods based on the PHD filter have successfully improved the tracking performance by using the detector fusion, appearance learning, and motion models.
% PHD-GM. GLMB-IM, MTDF 설명에 포함
\subsection{State-of-the-Art MOTS Methods}
% MOT to MOTS
Conventional MOT methods~\cite{gmphdn1tr, mtdf, fu1, fu2, phdgm, gmphd2012, eamtt, gmphdkcf, gmphdntype, gmphdogm, ycyoon1,ycyoon2,kjyoon1,kjyoon2,cbmot1,cbmot2} have exploited the tracking-by-detection paradigm, where the detectors~\cite{dpm,frcnn,yolo,sdp,regionlets} generate 2D bounding box results and the trackers assign tracking identities (IDs) to the bounding boxes via data association. Unlike MOT, MOTS uses pixelwise instance segmentation results as a tracking input instead of 2D bounding box results. P. Voigtlaender~\textit{et al.}~\cite{mots_trackrcnn} first introduced the MOTS task. They extended the popular MOT datasets such as KITTI~\cite{kitti} and MOTChallenge~\cite{mot16} with instance segmentation results by using a fine-tuned MaskRCNN~\cite{maskrcnn} for the same image sequences, and proposed a new soft multi-object tracking and segmentation accuracy (sMOTSA) measure that can be used to evaluate MOTS methods. In addition, they presented a new MOTS baseline method named TrackRCNN, which was extended from MaskRCNN with 3D convolutions to deal with temporal information.

% seg
Inspired by the new task, state-of-the-art MOTS methods~\cite{motsnet,MOTSFusionJ,PointTrack} have been proposed very recently. 
MOTSNet~\cite{motsnet} presents an intensive and semiautomated learning strategy for harvested datasets, e.g., ImageNet~\cite{imagenet} and Mapillary Vistas~\cite{mapillary} to improve instance segmentation quality. Additionally, these methods use a novel mask-pooling layer for improved object association over time. MOTSFusion~\cite{MOTSFusionJ} proposes a fusion-based MOTS method exploiting bounding box detection~\cite{rrc} and instance segmentation~\cite{bb2segnet}. It estimates a segmentation mask for each bounding box and builds up short tracklets using 2D optical flow, then fuses these 2D short tracklets into dynamic 3D object reconstructions hierarchically. The precise reconstructed 3D object motion is used to recover missed objects with occlusions in 2D coordinates.
PointTrack~\cite{PointTrack} devises a new feature extractor based on PointNet~\cite{PointNet} to appropriately consider both foreground and background features. This is motivated by the fact that the inherent receptive field of convolution-based feature extraction inevitably confuses up the foreground and background features. PointNet is used to randomly sample feature points considering the offsets between foreground and background regions, the colors of those regions, and the categories of segments. Then the context-aware embedding vectors for association are built after concatenation of the separately computed position difference vectors.

Motivated by these state-of-the-art MOTS studies~\cite{mots_trackrcnn,motsnet,MOTSFusionJ,PointTrack}, we propose a highly practical MOTS method without intensive or additional learning in~\cite{motsnet, PointNet} for more precise pixelwise segmentation and without requiring multi-domain data such as detection, segmentation, and 2D-to-3D reconstruction to perform fusion~\cite{MOTSFusionJ}.
Our proposed method, named GMPHD\_MAF, exploits the tracking-by-instance-segmentation paradigm, which uses only 2D images and one instance segmentation result as inputs and performs the MOTS task by using two popular filtering methods, the GMPHD filter~\cite{Vo_gmphd} and the KCF~\cite{kcf}, with simple parameter tuning. We build a two-step hierarchical data association strategy to handle tracklet loss and ID switches. In each association stage, position and motion affinity are calculated by the GMPHD filter, and appearance affinity is calculated by the KCF. To appropriately consider these two affinities, we propose a mask-based affinity fusion model. GMPHD\_MAF shows comparative performance in two popular KITTI-MOTS~\cite{kitti} and MOTSChallenge~\cite{mot16} datasets against state-of-the-art MOTS methods~\cite{mots_trackrcnn,motsnet,MOTSFusionJ,PointTrack}.

% 여러 방법들이 학습을 통해 segmentation 고도화 하고 fusion 하는 정보를 늘리고, scenc context 및 random sample을 이용해서 segmentation 고도화하고 객체들 간의 disriminative 를 잘 구별하는 vector를 뽑고 하는 방법을 시도헀지만 heavy 하다 실용적이지 못하다. 그래서 우리는.. 앞에 어떤 방법이 나오든 바로 적용가능한 실용적이고 직관적인 online 방법을 제시한다.
\begin{figure}[t]
\begin{center}
   \includegraphics[width=8.5cm]{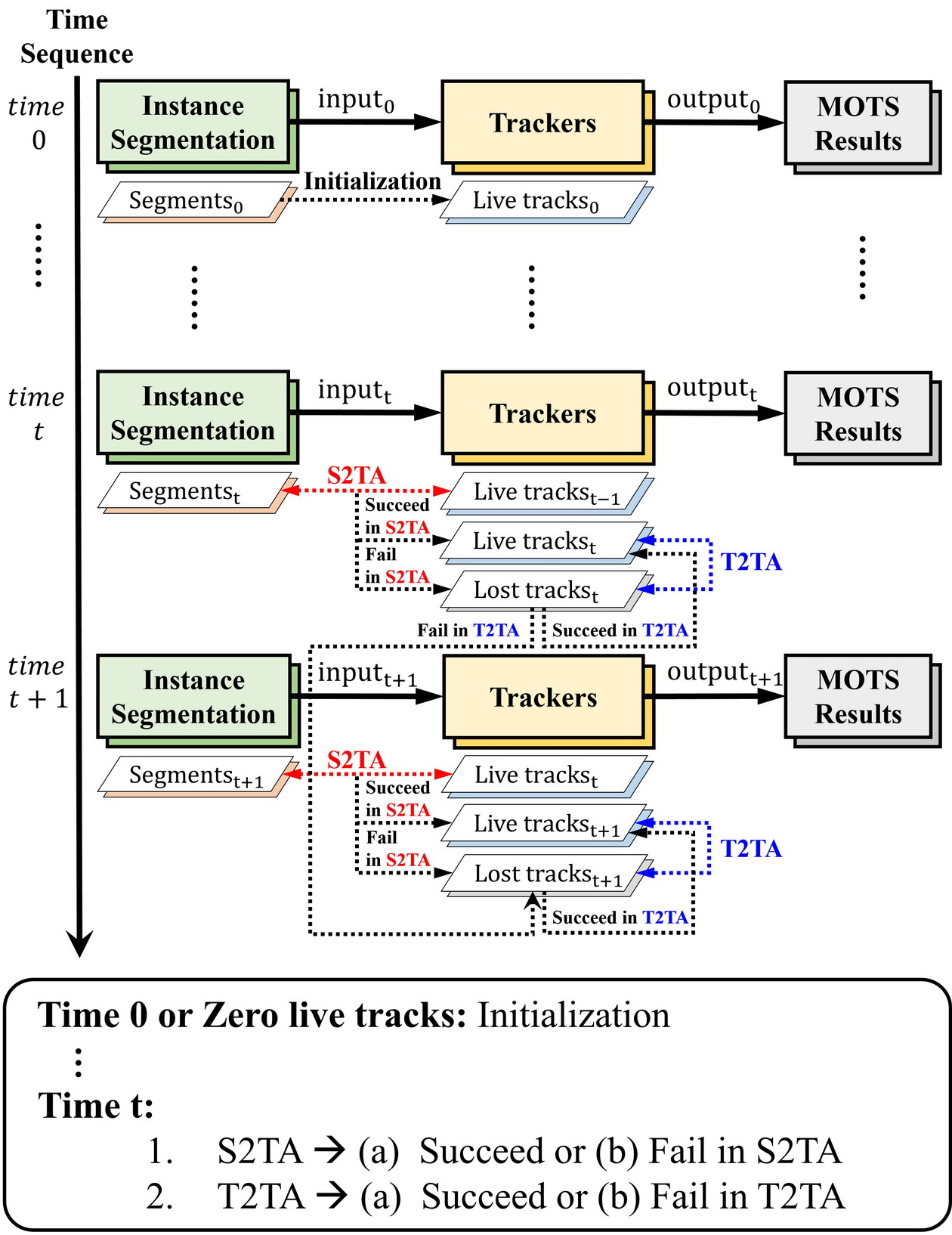}
\end{center}
   \caption{Processing pipeline of the proposed online multi-object tracking and segmentation framework, where segment-to-track association and track-to-track association are abbreviated as S2TA and T2TA, respectively. As shown in this figure, the fully online approach uses only past and current information, with no intentional frame latency.}
   % \caption{Examples of detection results: (a) 2D bounding boxes, (b) 3D boxes, and (c) instance segmentation in KITTI dataset. The results are visualized in the same image but (a) and (c) were obtained from the camera images and (b) was obtained from Velodyne LiDAR 3D point clouds and calibrated to the camera image coordinates.}
\label{fig2}
\end{figure}

\begin{figure*}
\begin{center}
\includegraphics[width=18cm]{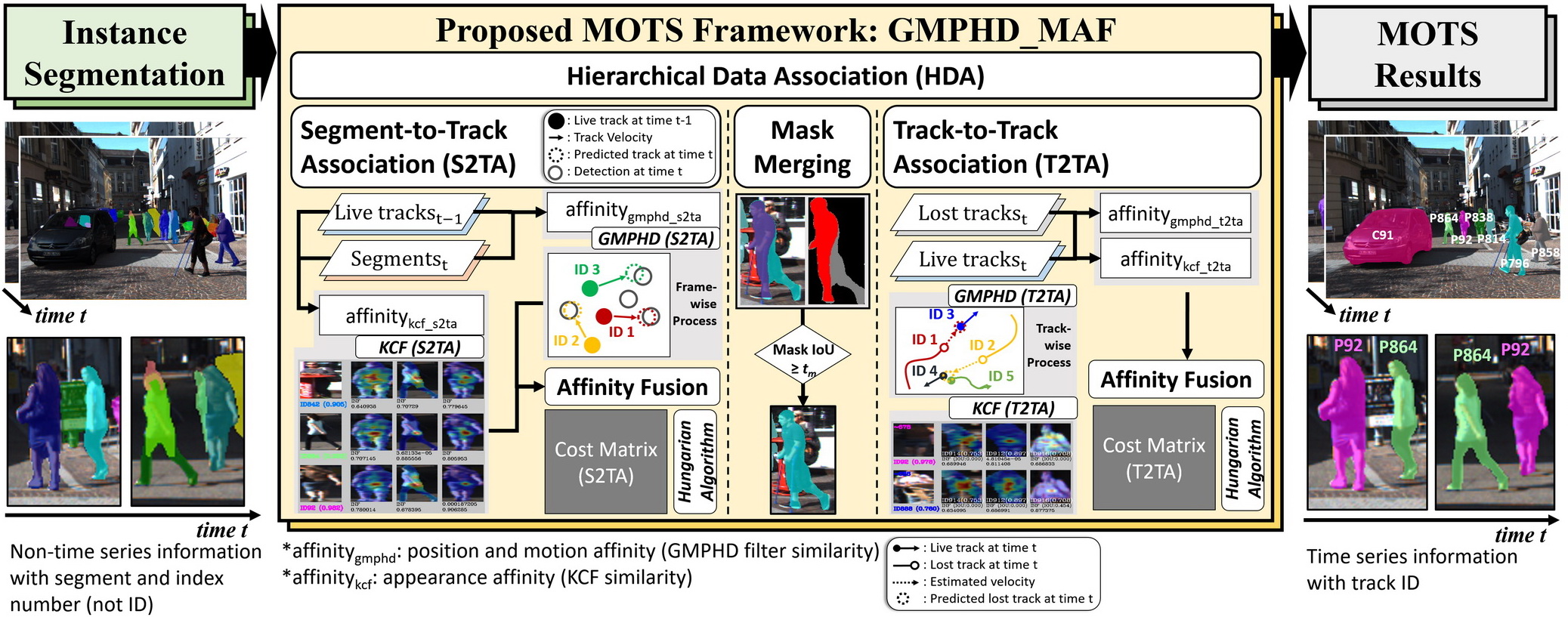}
\end{center}
   \caption{Detailed processing pipeline of GMPHD\_MAF with input (images and instance segmentation results) and output (MOTS results). The key components are hierarchical data association (HDA), mask merging, and mask-based affinity fusion (MAF). HDA has two association steps: S2TA and T2TA. MAF executes each affinity fusion in each association step while mask merging runs once between S2TA and T2TA.}
\label{fig3}
\end{figure*}

\section{Proposed Method}
\label{sec:proposed}
In this section, we introduce the proposed online multi-object tracking and segmentation (MOTS) framework in terms of input/output interfaces (I/O) and key modules in detail.
Following the tracking-by-segmentation paradigm, the MOTS method receives image sequences and instance segmentation results as inputs and gives MOTS results as outputs, which are shown in Figures~\ref{fig1} and~\ref{fig3}. Each instance has an object type, pixelwise segment, and confidence score but does not include time series information. Through the MOTS method, we can assign tracking IDs to the object segments and turn them they turn into time series information, i.e., MOTS results.

The proposed MOTS framework is not only built based on a HDA strategy consisting of segment-to-track association (S2TA) and track-to-track association (T2TA) but is also implemented as a fully online process using only information at the present time $t$ and the past times $0$ to $t-1$ (see Figure~\ref{fig2}).
In each observation-to-state association stage, affinities between states and observations are calculated considering position, motion, and appearance. The ``position and motion" and ``appearance" affinities are computed by using a GMPHD filter~\cite{Vo_gmphd} and a KCF~\cite{kcf}, respectively. Since these two types of affinities have different filtering domains, one affinity can be of a much higher magnitude than the other affinity. To appropriately consider position, motion, and appearance information in HDA, we devise a MAF method. Additionally, to reduce false positive segments, we adopt the mask intersection-over-union (IoU)-based merging technique between S2TA and T2TA.

In summary, the proposed MOTS framework follows the order of (1) S2TA, (2) mask merging, and then (3) T2TA, in which the affinities of each association are computed by exploiting the GMPHD filter and KCF, are fused by using MAF.
In what follows, we use GMPHD\_MAF as the abbreviation for the proposed framework (see Figure~\ref{fig3}).

% pseudocode in Algorithm~\ref{algo}.

\subsection{GMPHD Filtering Theory}
\label{sub:gmphd}
The main steps of the GMPHD filtering-based tracking includes initialization, prediction, and update.
% 자세한 활용예시는 섹션~~에 있다. 여기서는 우리방법의 이해를 위한 이론과 notation terminolgy 를 설명한다.
The set of states (segment tracks) and the set of observations (instance segmentations) at time $t$ are $X_t$ and $Z_t$ represented as follows:
%{\footnotesize
\begin{align}
&X_t = \{\mathbf{x}^1_t,\dots,\mathbf{x}^{N_{t}}_t\}, \label{eq:state}\\
&Z_t = \{\mathbf{z}^1_t,\dots,\mathbf{z}^{M_{t}}_t\}, \label{eq:obs}
\end{align}
%}%
where a state vector $\mathbf{x}_t$ is composed of \{$c_x$, $c_y$, $v_x$, $v_y$\} with a tracking ID, and segment mask. $c_x$, $c_y$, and $v_x$, $v_y$ indicate the center coordinates of the mask's 2D box, and the velocities of the $x$ and $y$ directions of the object, respectively. An observation vector $\mathbf{z_t}$ is composed of \{$c_x$, $c_y$\} with a segment mask. The Gaussian model $\mathcal{N}$ representing $\mathbf{x_t}$ is initialized by $\mathbf{z_t}$, predicted to $\mathbf{x_{t+1|t}}$, and updated to $\mathbf{x_{t+1}}$ by $\mathbf{z_{t+1}}$.

\textit{\textbf{Initialization}}:
The Gaussian mixture model $g_t$ are initialized by using the initial observations from the detection responses. In addition, when an observation fails to find the association pair, i.e., to update the target state, the observation initializes a new Gaussian model. We call this \textit{birth} (a kind of initialization). Each Gaussian $\mathcal{N}$ represents a state model with weight $w$, mean vector $m$, input state vector $\mathbf{x}$, and covariance matrix $P$, which are as follows: 
\begin{align}
    g_t(\mathbf{x})=\sum _{ i=1 }^{ { N }_{ t } }{ { w }_{ t }^{ i }\mathcal{N}(\mathbf{x};\mathbf{ m }_{ t }^{ i },{ P }_{ t }^{ i }) }, \label{eq:gmm}
\end{align}
where $N_t$ is the number of Gaussian models. At this step, we set the initial velocities of the mean vector to zero. Each weight is set to the normalized confidence value of the corresponding detection response. Additionally, the method of setting covariance matrix $P$ is shown in Section~\ref{subsec:params}.

\textit{\textbf{Prediction}}: We assume that there already has been the Gaussian mixture $g_{t-1}$ of the target states at the previous frame $t-1$, as shown in~\eqref{eq4}. Then, we can predict the state at time $t$ using Kalman filtering. In~\eqref{eq5}, $\mathbf{m}_{t|t-1}^i$ is derived by using the velocity at time $t-1$ and the covariance $P$ is also predicted by the Kalman filtering method in~\eqref{eq6} as:
\begin{align}
    &g_{t-1}(x)=\sum _{ i=1 }^{ { N }_{ t-1 } }{ { w }_{ t-1 }^{ i }\mathcal{N}(\mathbf{x};\mathbf{ m }_{ t-1 }^{ i },{ P }_{ t-1 }^{ i }) }, \label{eq4}\\
    &\mathbf{ m }_{ t|t-1 }^{ i }={ F }\mathbf{ m }_{ t-1 }^{ i },\label{eq5}\\
    &{ P }_{ t|t-1 }^{ i }={ Q }+{ F }{ P }_{ t-1 }^{ i }{ { (F }) }^{ T },\label{eq6}
\end{align}
where $F$ is the state transition matrix, and $Q$ is the process noise covariance matrix. Those two matrices are constant in our tracker.

\textit{\textbf{Update}}: The goal of the update step is to derive~\eqref{eq7}. First, we should find an optimal observation $\mathbf{z}$ at time $t$ to update the Gaussian model. The optimal $\mathbf{z}$ in the observation set $Z$ makes $q_t$ the maximum value in~\eqref{eq8} as:
\begin{align}
 &g_{t|t}(\mathbf{x})=\sum _{ i=1 }^{ { N }_{ t|t } }{ { w }_{ t }^{ i }(\mathbf{z})\mathcal{N}(\mathbf{x};\mathbf{ m }_{ t|t }^{ i },{ P }_{ t|t }^{ i }) }, \label{eq7}\\
 &{q}_{t}^{i}(\mathbf{z})=\mathcal{N}(\mathbf{z};{H}\mathbf{m}_{t|t-1}^{i},{R}+{H}{P}_{t|t-1}^{i}{({H})}^{T}).\label{eq8}
 \end{align}
 From the perspective of application, the update step involves data association. Finding the optimal observations and updating the state models is equivalent to finding the association pairs. $R$ is the observation noise covariance. $H$ is the observation matrix uesd to transform a state vector into an observation vector. Both matrices are constant in our application. After finding the optimal $\mathbf{z}$, the Gaussian mixture is updated in the order of~\eqref{eq9}, \eqref{eq10}, \eqref{eq11}, and \eqref{eq12} as:
 \begin{align}
 &{w}_{t}^{i}(\mathbf{z})=\frac {{w}_{t|t-1}^{i}{q}_{t}^{i}(\mathbf{z})}{\sum _{l=1}^{{N}_{t|t-1}}{{w}_{t|t-1}^{l}{q}_{t}^{l}(\mathbf{z})}  }, \label{eq9}\\
&\mathbf{ m }_{ t|t }^{ i }(\mathbf{z})=\mathbf{ m }_{ t|t-1 }^{ i }+{ K }_{ t }^{ i }{ (\mathbf{z}-{ H }\mathbf{ m }_{ t|t-1 }^{ i }) },\label{eq10}\\
 &P_{t|t}^i=[I-{K_t}^{i}{H}]{P_{t|t-1}^i},\label{eq11}\\
 &K_t^i=P_{t|t-1}^i{H}^T({H}P_{t|t-1}^i{H}^T+R)^{-1},\label{eq12}
\end{align}
where the set of ${w}_{t|t-1}$ includes ${w}_{t-1}$ (weights from the targets at the previous frame) and ${w}_{t-1}$ (weights of newly born targets). Likewise, ${N}_{t|t-1}$ is the sum of ${N}_{t-1}$ and the number of the newly born targets.

\subsection{Hierarchical Data Association (HDA)}

\begin{figure}[t]
\begin{center}
\includegraphics[width=8.5cm]{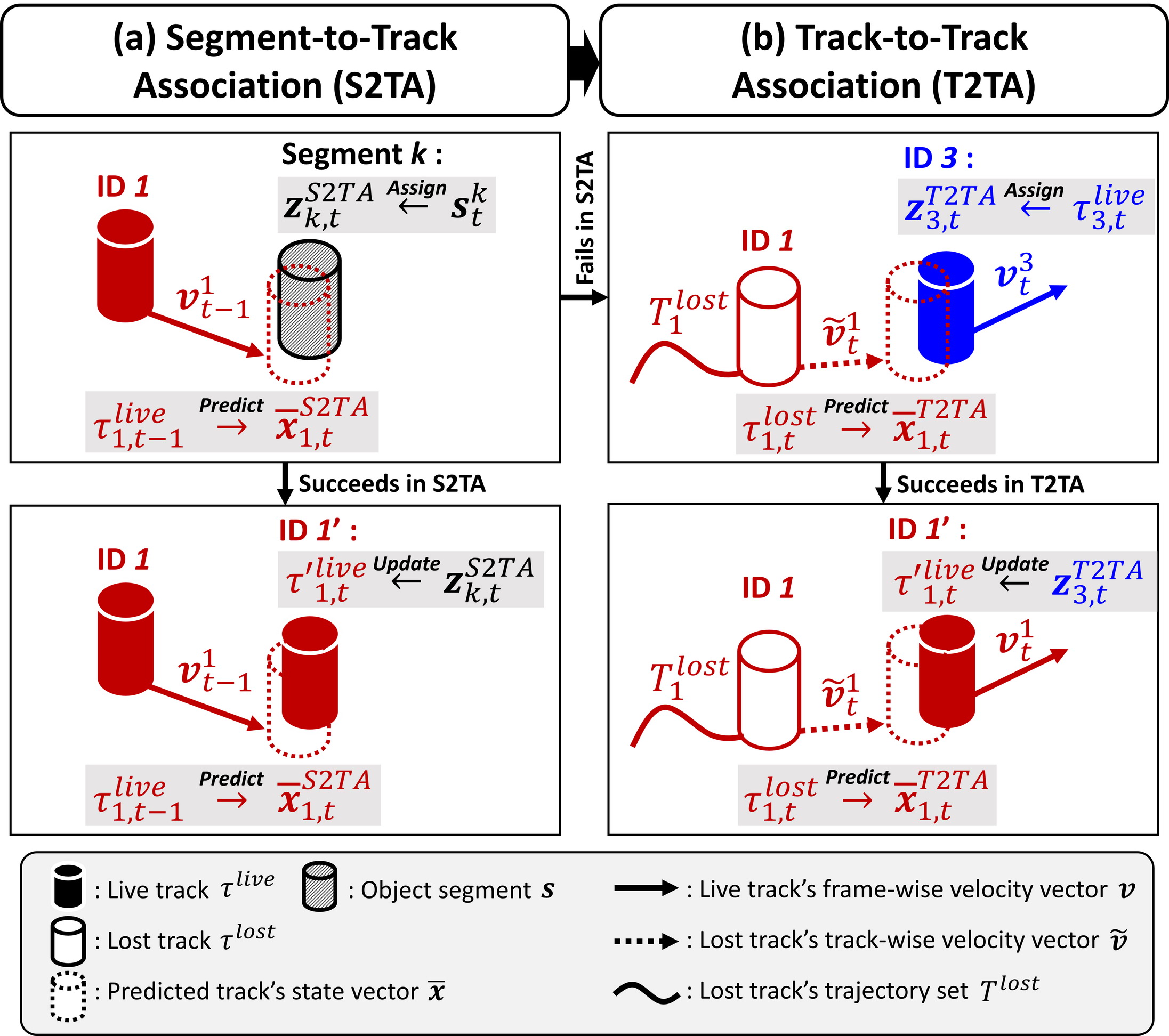}
\end{center}
   \caption{Detailed example of the GMPHD filter-based hierarchical data association (HDA) process with (a) S2TA and (b) T2TA. Track $\tau^{live}_{1,t-1}$ can be associated with an observation $\mathbf{z}^{S2TA}_{k,t}$ from segment $k$. If S2TA fails, $\tau^{live}_{1,t_1}$ becomes $\tau^{lost}_{1,t}$, and then it can be associated with an observation $\mathbf{z}^{T2TA}_{3,t}$ from the live track $\tau^{live}_{3,t}$.}
\label{fig4}
\end{figure}

\begin{figure}[t]
\begin{center}
\includegraphics[width=8.5cm]{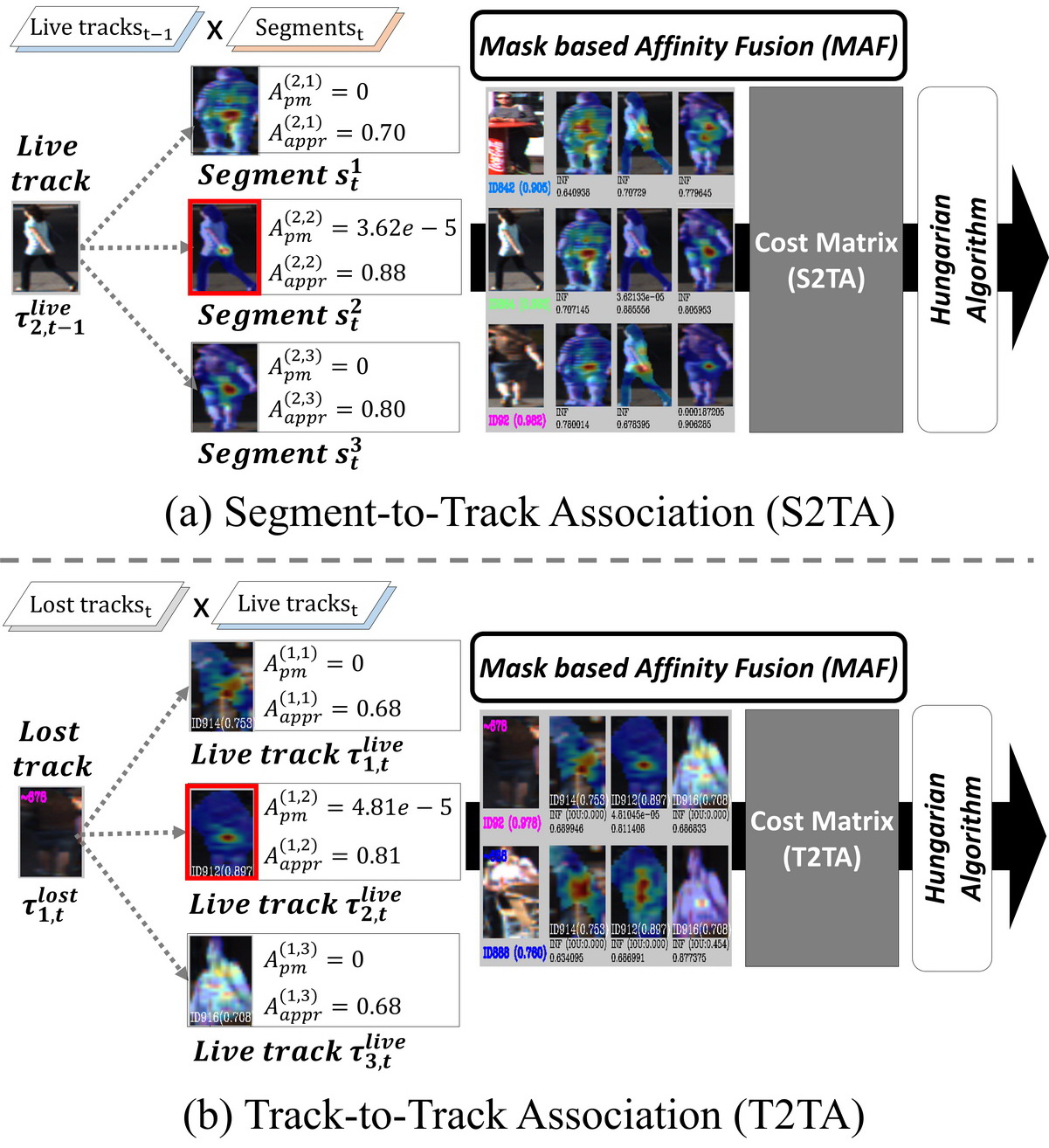}
\end{center}
   \caption{Example of the proposed mask-based affinity fusion (MAF) method in (a) S2TA and (b) T2TA.}
\label{fig5}
\end{figure}

To compensate for the imperfection of the framewise one-step online propagation of the GMPHD filtering process, we extend the GMPHD filter-based online MOT with a hierarchical data association (HDA) strategy that has two-step association steps: S2TA and T2TA (see Figure~\ref{fig4}). Each association has different states and observations as inputs, which are used to compute position and motion affinity$_{pm}$ and appearance affinity$_{appr}$ (see Figure~\ref{fig5}).

To build the proposed HDA strategy, we define some online MOT's processing units at the present time $t$. $S_t$ indicates the instance segmentation results and the $k^{th}$ segment is denoted by $s^{k}_{t}$. $T$ indicates a set of tracks. These units are defined in detail as:
\begin{align}
&S_t = \{s^{1}_{t},...,s^{k}_{t}\},\label{eq13}\\
&T^{live}_t = \{\tau^{live}_{1,{t}},...,\tau^{live}_{i,{t}}\},\label{eq14}\\
&\tau^{live}_{i,t} = \{\mathbf{x}^i_{t_{b}},..,\mathbf{x}^i_{t_{l}}\},\quad\quad\,\,\,\, 0\le t_{b}<t_{l},\,t_{l}=t,\label{eq15}\\
&T^{lost}_t = \{\tau^{lost}_{1,t},...,\tau^{lost}_{j,t}\},\label{eq16}\\
&\tau^{lost}_{j,t} = \{\mathbf{x}^j_{t_{b}},..,\mathbf{x}^j_{t_{l}}\},\quad\quad\,\,\,\, 0\le t_{b}<t_{l}<t,\label{eq17}\\
&\mathbf{x}_{t_{b}} = \{c_{x,{t_{b}}},c_{y,{t_{b}}},v_{x,{t_{b}}},v_{y,{t_{b}}}\}^T,\label{eq18}\\
&\mathbf{x}_{t_{l}} = \{c_{x,{t_{l}}},c_{y,{t_{l}}},v_{x,{t_{l}}},v_{y,{t_{l}}}\}^T,\label{eq19}
\end{align}
where \textit{``live"} indicates that tracking succeeds at time $t$. \textit{``lost"} indicates that tracking fails at time $t$. The two attributes are not compatible, and $T^{lost}_t \cup T^{live}_t = T^{all}_t$ and $T^{lost}_t \cap T^{live}_t = \phi$ are satisfied.
$T_t$ is composed of a track $\tau_{i,t}$ with identity $i$ which is also a set of state vectors from the birth time ${t_b}$ to the last tracking time ${t_l}$. In the case of $\tau^{live}_{i,t}$, ${t_l}$ is identical to the present time $t$, in the case of $\tau^{lost}_{j,t}$, ${t_l}$ is less than time $t$. Regardless of when time $t$ is, state vector $\mathbf{x}$ has the center point $\{c_x, c_y\}$ in the segment bounding box, velocities $\{v_x, v_y\}$ in the directions of the $x$-axis and y-axis, an identity (ID), and a segment mask (see \eqref{eq18} and \eqref{eq19}). 

\noindent{\textbf{Segment-to-track association (S2TA):}}
In S2TA, the observations denoted by $Z^{S2TA}_{t}$ are frame-by-frame instance segmentation results $S_{t}$. If there are no track states, the states $X^{S2TA}_{t}$ are initialized from $Z^{S2TA}_{t}$, and otherwise, $\bar{X}^{S2TA}_{t}$ is predicted from $T^{live}_{t-1}$ and updated by using the GMPHD filter with the processing units as follows:
\begin{align}
&\mathbf{z}^{S2TA}_{k,t} = \{c^{k}_{x,t},c^{k}_{y,t}\}^T \text{from } s^k_t,\label{eq20}\\
&\mathbf{x}^{S2TA}_{i,t-1} = \{c^i_{x,t-1},c^i_{y,t-1},v^i_{x,t-1},v^i_{y,t-1}\}^T \text{from } \tau^{live}_{i,{t-1}},\label{eq21}\\
&F^{S2TA} = \begin{pmatrix} 1 & 0 & 1 & 0  \\ 0 & 1 & 0 & 1  \\ 0 & 0 & 1 & 0  \\ 0 & 0 & 0 & 1  \end{pmatrix},\label{eq22}\\
&\mathbf{\bar{x}}^{S2TA}_{i,t} = F^{S2TA}\mathbf{x}^{S2TA}_{i,{t-1}},\label{eq23}\\
&\mathbf{\bar{x}}^{S2TA}_{i,t} = \{\bar{c}^i_{x,t},\bar{c}^i_{y,t},v^i_{x,t-1},v^i_{y,t-1}\}^T,\label{eq24}
\end{align}
where~\eqref{eq22} is identical to~\eqref{eq5} from the~\textit{Prediction} step and $\bar{c}^i_t$ is equal to ${c}^i_{t-1}+{v}^i_{t-1}$. More details can be seen in Figure~\ref{fig4}.

Depending on whether the state $x$ finds that an observation $z$ is associated, is born, or neither, the framewise motion $v$ is updated as follows:
{\footnotesize
\begin{align}
    \mathbf{v}^i_{t}= 
\begin{cases}
    \beta*\mathbf{v}^i_{t-1}+(1.0-\beta)*\begin{Bmatrix}
c^{k}_{x,t}-\bar{c}^{i}_{x,t}\\ 
c^{k}_{y,t}-\bar{c}^{i}_{y,t}
\end{Bmatrix}
, & 
    \text{if } \mathbf{z}^k_t \text{ is assigned to }\mathbf{\bar{x}}^{S2TA}_{i,t}\\
    \{0,0\}^T,              & \text{else if } \mathbf{x}^{S2TA}_{i,t}\text{ is born} \\
    \mathbf{v}^i_{t-1},              & \text{otherwise}
\end{cases},\label{eq25}
\end{align}
}%
where $\beta$ can be set differently set according to the scene context and frame rate.

%%%%%%%%%%%%%%%%%%%%%%%%%%%%%%%%%%%%%%%%%%%%%%%%%%%%%%%%%%
%\begin{figure}[t]
%\begin{center}
%\includegraphics[width=8.5cm]{figures/fig6_2_resize.jpg}
%\end{center}
%   \caption{Exemplar of the proposed mask based affinity fusion (MAF) method in (a) S2TA and (b) T2TA.}
%\label{fig5}
%\end{figure}
%%%%%%%%%%%%%%%%%%%%%%%%%%%%%%%%%%%%%%%%%%%%%%%%%%%%%%%%%%

\noindent{\textbf{Track-to-track association (T2TA):}}
In T2TA, observations $Z^{T2TA}_{t}$ and states $X^{T2TA}_{t}$ (inputs) are built from the live track set $T^{live}_t$ and lost track set $T^{lost}_t$, respectively. Each of $T^{live}_t$ and $T^{lost}_t$ consists of the track vectors of $\tau^{live}_{i,t}$ and $\tau^{lost}_{j,t}$ with their identities (see \eqref{eq14} and \eqref{eq16}). The track vectors have temporal information with the birth time ${t_b}$ and loss time ${t_l}$. The live track's ${t_l}$ is identical to the current time $t$, which means that the track is not yet lost, while the lost track's ${t_l}$ is less than $t$, which means the track was lost before the time $t$  (see \eqref{eq15} and \eqref{eq17}). 

Unlike the~\textit{Prediction} step of S2TA, where the framewise motion from time $t-1$ to $t$ is used, a trackwise motion analysis is used in T2TA as follows:
\begin{align}
\mathbf{z}^{T2TA}_{i,t} &= \{c^{i}_{x,t},c^{i}_{y,t}\}^T \text{from } \tau^{live}_{i,{t}},\label{eq26}\\
\mathbf{x}^{T2TA}_{j,{t-1}} &= \{c^j_{x,t_l},c^j_{y,t_l},\tilde{v}^j_{x,t},\tilde{v}^j_{y,t}\}^T \text{from } \tau^{lost}_{j,{t}},\label{eq27}\\
F^{T2TA} &= \begin{pmatrix} 1 & 0 & d_f & 0  \\ 0 & 1 & 0 & d_f  \\ 0 & 0 & 1 & 0  \\ 0 & 0 & 0 & 1  \end{pmatrix},\label{eq28}\\
\mathbf{\bar{x}}^{T2TA}_{j,t} &= F^{T2TA}\mathbf{x}^{T2TA}_{j,{t-1}},\label{eq29}\\
\mathbf{\bar{x}}^{T2TA}_{j,t} &= \{\bar{c}^j_{x,t},\bar{c}^j_{y,t},v^j_{x,t},v^j_{y,t}\}^T,\label{eq30}
\end{align}
where $d_{f}(i,j)$~\eqref{eq28} is the frame difference between $\tau^{live}_{i,t}$'s first element $\mathbf{x}^i_{t_b}$~\eqref{eq15} and $\tau^{lost}_{j,t}$'s last element $\mathbf{x}^j_{t_l}$~\eqref{eq17}. The trackwise motion vector $\{\tilde{v}^j_{x,t},\tilde{v}^j_{y,t}\}^T$ of~\eqref{eq27} is calculated as follows:
\begin{align}
\mathbf{\tilde{v}^j_t} = \{\tilde{v}^j_{x,t},\tilde{v}^j_{y,t}\}^T = \{\frac{c^j_{x,t_l}-c^j_{x,t_b}}{t_l-t_b},\frac{c^j_{y,t_l}-c^j_{y,t_b}}{t_l-t_b}\}^T,\label{eq31}
\end{align}
where $\frac{c^j_{x,t_l}-c^j_{x,t_b}}{t_l-t_b},\frac{c^j_{y,t_l}-c^j_{y,t_b}}{t_l-t_b}$ are the averaged velocities in the directions of the x-axis and y-axis, respectively. The velocities are computed by subtracting the center position of the first object state $\mathbf{x}^j_{t_b}$ from that of the last state $\mathbf{x}^j_{t_l}$ and dividing it by the frame difference $t_l - t_b$, which is equivalent to the length of the track $\tau^{lost}_{j,t}$. 

In terms of temporal motion analysis, 
S2TA has the same time interval \textit{``1"} between states and observations in transition matrix $F$, whereas T2TA has a different time interval (frame difference) between states and observations. The variable $d_f$ depends on which state of the lost track and observation of the live track are paired. \eqref{eq20}-\eqref{eq25} of S2TA are the prediction step with framewise motion analysis and update, but \eqref{eq26}-\eqref{eq31} of T2TA contain the prediction step with trackwise linear motion analysis. A detailed example is shown in Figure~\ref{fig4}.

Following the proposed HDA strategy, for S2TA and T2TA, two cost matrices can be filled by using the affinities between the differently defined states and observations. In the next subsection, we present an efficient mask-based affinity calculation method considering position, motion, and appearance for multi-object tracking and segmentation.

\subsection{Mask-Based Affinity Fusion (MAF)}
\label{sub:maf}

We adopt a simple score-level fusion method to adequately consider position, motion, and appearance between states and observations.
Fusing affinities obtained from different domains requires a normalization step that can balance the different affinities and avoid bias toward one affinity, which may have a much higher magnitude than the others.

\noindent{\textbf{Position and motion affinity:}} The GMPHD filter includes Kalman filtering in its \textit{Prediction} step,~\eqref{eq4}-\eqref{eq6}, designed with a linear motion model with noise $Q$. Additionally, we present two different linear motion models for the hierarchical data association with two stages, S2TA and T2TA, as described in~\eqref{eq25} and ~\eqref{eq31}. Therefore, the position and motion affinity between the $i^{th}$ state and $j^{th}$ observation gives the probabilistic value $w \cdot q(\mathbf{z})$ obtained by the GMPHD filter as follows:

%
%{\footnotesize
\begin{align}
    A^{(i,j)}_{pm} = {{w}^{i}\cdot{q}^{i}(\mathbf{z}^j)},\label{eq:affinitypm}
\end{align}
%}%
which is acquired from ~\eqref{eq8} and~\eqref{eq9} of the~\textit{Update} step.
%of the GMPHD filter~\cite{gmphdsaf_suppl}.

% \noindent{\textbf{Appearance affinity.}} 
\noindent{\textbf{Appearance affinity:}} We exploit the KCF~\cite{kcf} to compute the appearance affinity between the $i^{th}$ state and $j^{th}$ observation since instance segmentation results does not provide appearance features to discriminate the objects belonging to a single class, pedestrian or cars. The KCF does not have class dependency because it was originally designed for single-object tracking challenges such as the VOT benchmark~\cite{vot15}. So it can be applied multi-class tracking and we utilize it for calculating the appearance similarity by matching object templates in this paper. 
Before applying the KCF, the state and observation image templates are preprocessed by setting the backgrounds pixels to zero in the RGB channel's 0 to 255 ranges. This preprocessing step ensures that he appearance affinity pays attention to the foreground pixels based on the segment mask. The KCF-based affinity can be derived as follows: 
%
%{\footnotesize
\begin{align}
   A^{(i,j)}_{appr} = 1 - \frac { \sum _{ c=x_j }^{ width_j }{ \sum _{ r=y_j }^{ height_j }{ \bar{d}^{(i,j)}_{ KCF }(r,c) }  }  }{ width_j\cdot height_j } ,\label{eq:affinityappr}
\end{align}
%}%
where $\bar{d}(\cdot)$ indicates the normalized KCF distance value, which varies from $0.0$ to $1.0$ at each pixel. 
More intensive and advanced single-object tracking and template matching methods such as SiamRPN~\cite{siamrpn} and reidentification~\cite{reidsurvey} can be adopted in computing this affinity but it is beyond the scope of this paper.

\noindent{\textbf{Min-max normalization:}} In our experiments, $A_{pm}$ and $A_{appr}$ have quite different magnitudes, e.g.,  $A_{pm}=\{10^{-9},\dots,10^{-3}\}$ and $A_{appr}=\{0.4,\dots,1.0\}$ (see Figures~\ref{fig5}-\ref{fig7}). To fuse two affinities, we apply min-max normalization to them as follows:

%
%{\footnotesize
\begin{align}
\bar{A}^{(i,j)}=\frac{{A}^{(i,j)}-\min_{\substack{1\le i\le N\\1\le j\le M}}{{A}^{(i,j)}}}{\max_{\substack{1\le i\le N\\1\le j\le M}}{ A^{(i,j)}} -\min_{\substack{1\le i\le N\\1\le j\le M}}{A^{(i,j)}}}.\label{eq:mmnorm}
\end{align}
%}%

Then, we finally propose a MAF model represented by:

%
%{\footnotesize
\begin{align}
    {A}^{(i,j)}_{maf}=\bar{A}^{(i,j)}_{pm}\bar{A}^{(i,j)}_{appr}.\label{eq:maf}
\end{align} 
From this fused affinity, we can compute the final cost between states and observations as follows:
\begin{align}
    Cost(\mathbf{x^i_{t|t-1}},\mathbf{z_t^j})=-\alpha\cdot\ln{{A}^{(i,j)}_{maf}},\label{eq:maf_cost}
\end{align}
%}%
where $\alpha$ is a scale factor empirically set to $100$. If one of the affinities is close to zero, such as $10^{-39}$, the cost is set to $10000$ to prevent the final cost from becoming an infinite value. Then, the final cost ranges from $0$ to $10000$.

From the different states and observations (inputs) in S2TA and T2TA, two cost matrices are computed in every frame and we utilize the Hungarian algorithm~\cite{hungarian} to solve the cost matrices, as shown in Figure~\ref{fig5}. Then, observations succeeding in S2TA or T2TA are assigned to the associated states for \textit{Update}, and other observations failing in S2TA and T2TA initialize new states.

\subsection{Mask Merging}
\label{subsec:merging}
As shown in Figure~\ref{fig3}, for mask merging, i.e., track merging, we can utilize bounding box-based IoU or segment mask-based IoU (mask IoU) measures that calculate boxwise or pixel-wise overlapping ratios between two objects, respectively.
The two measures are represented by:
%
%{\footnotesize
\begin{align}
\label{eq:iou}
&\text{IoU}(A,B) = \frac{\text{bbox}(A) {\cap} \text{bbox}(B)}{\text{bbox}(A) {\cup} \text{bbox}(B)},\\
\label{eq:mask_iou}
&\text{Mask IoU}(A,B) = \frac{\text{mask}(A) {\cap} \text{mask}(B)}{\text{mask}(A) {\cup} \text{mask}(B)}.
\end{align}
%}% 

If the value of a selected measure is greater than or equal to the threshold $t_m$, the two objects are merged into one object. Mask merging is applied only between tracking objects, i.e., states, that are not observations, after S2TA.

\subsection{Parallel Processing}
% experiment 나 구현쪽으로 뺄수도 있다.
We assume that data association runs only between the same class of objects.
Thus, if the instance segmentation module provides two or more object classes, e.g., car and pedestrian classes,
our proposed framework is easily expansible (see Figure~\ref{fig1}). In this paper, we implement the MOTS module with two parallel MOTS processes because the datasets used for our experiments produce car and pedestrian segments. 

\section{Experiments}
\label{sec:experiments}
\begin{figure}[t]
\begin{center}
   \includegraphics[width=8.5cm]{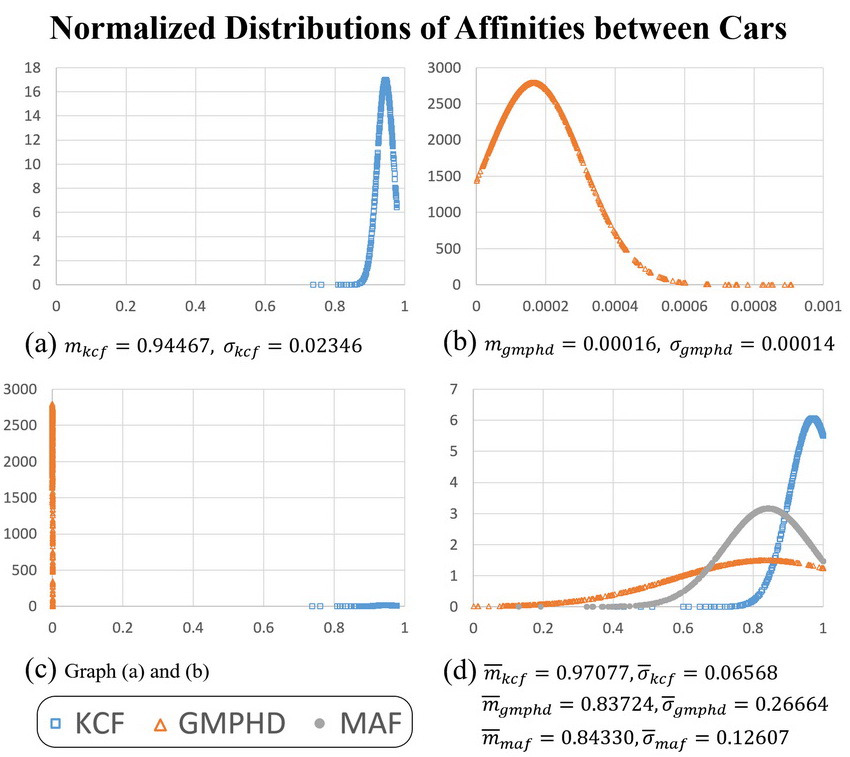}
\end{center}
   \caption{Normalized distributions of the affinities between cars in KITTI-MOTS training sequence 0019. KCF and GMPHD represent ``appearance affinity" and ``position and motion affinity", respectively. (a) and (b) show the distributions with each average $m$ and standard deviation $\sigma$, and (c) shows that (a) and (b) are very different from each other. (d) The proposed mask-based affinity fusion (MAF) method can determine the scale difference between the KCF and GMPHD affinities and then normalize the two affinities and fuse (multiply) them. $\bar{m}$ and $\bar{\sigma}$ denote the normalized values in~\eqref{eq:mmnorm}. }
\label{fig6}
\end{figure}

\begin{figure}[t]
\begin{center}
   \includegraphics[width=8.5cm]{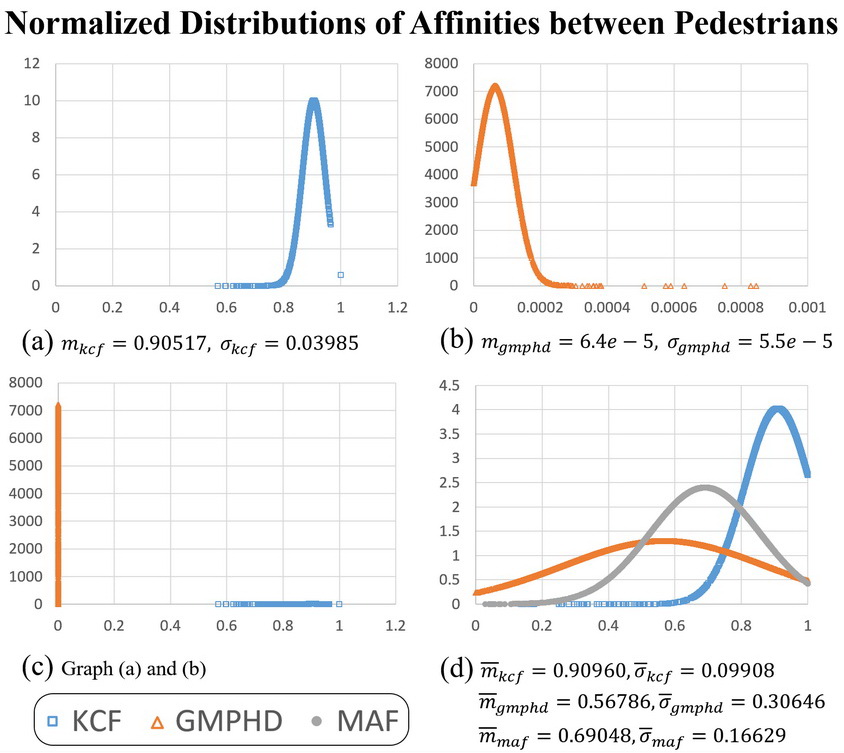}
\end{center}
   \caption{Normalized distributions of the affinities between pedestrians in KITTI-MOTS training sequence 0019. KCF and GMPHD represent ``appearance affinity" and ``position and motion affinity", respectively. (a) and (b) show the distributions with each average $m$ and standard deviation $\sigma$, and (c) shows that (a) and (b) are very different from each other. (d) The proposed mask-based affinity fusion (MAF) method can determine the scale difference between the KCF and GMPHD affinities and then normalize the two affinities and fuse (multiply) them. $\bar{m}$ and $\bar{\sigma}$ denote the normalized values in~\eqref{eq:mmnorm}. }
\label{fig7}
\end{figure}

In this section, we present experimental studies for the proposed MOTS metho, named GMPHD\_MAF, in detail. In \ref{sub:dataset} we note that GMPHD\_MAF is studied with state-of-the-art KITTI-MOTS~\cite{kitti} and MOTSChallenge~\cite{mots_trackrcnn} datasets and new evaluation measures. In \ref{sub:impl}, the implementation details of our method are addressed in terms of development environments and parameter settings. 
In~\ref{sub:ablation}, we determine the effectiveness of key modules through ablation studies in the dataset training sequences. The ablation studies show that the proposed key modules comprehensively improve the baseline model $p1$ remarkably in terms of IDS. In particular, the proposed MAF technique effectively fuses ``position and motion (GMPHD)" and ``appearance (KCF)" affinities, which are described in Figures~\ref{fig6} and~\ref{fig7}. Finally, in~\ref{sub:test}, we show that the final proposed model $p6$ achieves comparable performances on the test sequences of the datasets in terms of the sMOTSA, MOTSA, MOTSP, and IDS measures. 

\begin{table}[t]
\centering
\caption{\textbf{Evaluation measures}. sMOTSA has been mainly used for measuring the tracking performance as a key measure.}
\label{table:measures}
 \begin{tabular}{|c|c|c|p{4cm}|} 
 \hline
 \multicolumn{1}{|c|}{\footnotesize{Measure}} & \multicolumn{1}{c|}{\footnotesize{Better}} & \multicolumn{1}{c|}{\footnotesize{Perfect}} & \multicolumn{1}{c|}{\footnotesize{Description}} \\ \hline
 \footnotesize{MOTSA} & \footnotesize{$\uparrow$} & \footnotesize{$100 \%$} & 
\scriptsize{Multi-Object Tracking and Segmentation Accuracy~\cite{mots_trackrcnn}. This measure is the mask-based MOTS accuracy which combines four sources: TP, FN, FP, and IDS.}
\\ \hline
 \footnotesize{sMOTSA} & \footnotesize{$\uparrow$} & \footnotesize{$100 \%$} & 
\scriptsize{Soft Multi-Object Tracking and Segmentation Accuracy~\cite{mots_trackrcnn}. This measure is the soft mask-based MOTS accuracy which combines three sources: TP, FP, and IDS.}
\\ \hline
 \footnotesize{MOTSP} & \footnotesize{$\uparrow$} & \footnotesize{$100 \%$} & 
\scriptsize{Mask-overlap based variant of Multi-Object Tracking Precision~\cite{mots_trackrcnn}. The ratio of true positive masks among ground truth masks.}
\\ \hline
 \footnotesize{TP} & \footnotesize{$\downarrow$} & \footnotesize{0} & 
\scriptsize{Total number of true positive masks. }
\\ \hline
 \footnotesize{FP} & \footnotesize{$\downarrow$} & \footnotesize{0} & 
\scriptsize{Total number of false positive masks. }
\\ \hline
 \footnotesize{FN} & \footnotesize{$\downarrow$} & \footnotesize{0} & 
\scriptsize{Total number of false negative masks (missed targets).}
 \\ \hline
 \footnotesize{IDS} & \footnotesize{$\downarrow$} & \footnotesize{0} & 
\scriptsize{Total number of identity switches. Please note that we follow the stricter definition of identity switches as described in~\cite{ids}.}
\\ \hline
 \footnotesize{FPS} & \footnotesize{$\uparrow$} & \footnotesize{$\infty$} & 
\scriptsize{Processing speed (in frames per second excluding the detector) on the benchmark.}
\\ \hline
\end{tabular}
\end{table}
\subsection{Datasets and Measures}
\label{sub:dataset}
GMPHD\_MAF is evaluated on KITTI-MOTS~\cite{kitti} and MOTSChallenge~\cite{mots_trackrcnn}, which are the most popular datasets for MOTS.
P. Voigtlaender~\textit{et al.}~\cite{mots_trackrcnn} proposed new MOTS measures and two MOTS datasets that were extended from two representative MOTSChallenge~\cite{mot16} and KITTI~\cite{kitti} datasets.
They have been widely used for multi-object tracking with 2D bounding box-based detection results but instance segmentation results with the same image sequences were provided for MOTS, created by Mask R-CNN~\cite{maskrcnn} X152 of Detectron2~\cite{detectron2}.
For evaluation, sMOTSA and IDS are mainly used in this paper. 
These measures are mask-based variants of the original CLEAR MOT measures~\cite{clearmot} as follows:
\begin{align}
&\text{MOTSA} = \frac{|\text{TP}|-|\text{FP}|-|\text{IDS}|}{|\text{M}|},\label{eq:motsa}\\
&\widetilde{\text{TP}} =  \sum_{h \in \text{TP}}{\text{Mask IoU}(h,gt(h))} ,\label{eq:softtp}\\
&\text{sMOTSA} = \frac{\widetilde{\text{TP}}-|\text{FP}|-|\text{IDS}|}{|\text{M}|} ,\label{eq:smotsa}\\
&\text{MOTSP} = \frac{\widetilde{\text{TP}}}{|\text{TP}|},\label{eq:motsp}
\end{align}
where M is a set of ground truth (GT) pixel masks, $h$ is a track hypothesis mask, and $gt(h)$ is the most overlapping mask among all GTs. In multi-object tracking and segmentation accuracy (MOTSA), a mask-based variant of the original multi-object tracking accuracy (MOTA), a case is only counted as a true positive (TP) when the mask IoU value, between $h$ and $gt(h)$, is greater than or equal to $0.5$, but in soft multi-object tracking and segmentation accuracy (MOTSA), $\widetilde{\text{TP}}$ is used, which is a soft version of TP. Other details of the measures are displayed in Table~\ref{table:measures}.

\subsection{Implementation Details}
\label{sub:impl}
\noindent{\textbf{Development environments:}} 
All experiments are conducted on an Intel i7-7700K CPU @ 4.20GHz and DDR4 32.0GB RAM without GPU acceleration.
We implement GMPHD\_MAF by using OpenCV image processing libraries written in Visual C++. 
The code implementation is available at~\href{https://github.com/SonginCV/GMPHD_MAF}{\underline{https://github.com/SonginCV/GMPHD\_MAF}}.

\noindent{\textbf{Parameter settings:}}
\label{subsec:params}
The matrices F, Q, P, R, and H are used in \textit{Initialization}, \textit{Prediction}, and  \textit{Update}.
Experimentally, the parameter matrices for the GMPHD filter's tracking process are set as follows:

\begin{equation*}
    F = \begin{pmatrix} 1 & 0 & 1 & 0  \\ 0 & 1 & 0 & 1  \\ 0 & 0 & 1 & 0  \\ 0 & 0 & 0 & 1  \end{pmatrix},
    Q = \frac { 1 }{ 2 } \begin{pmatrix} 5^{ 2 } & 0 & 0 & 0  \\ 0 & 10^{ 2 } & 0 & 0  \\ 0 & 0 & 5^{ 2 } & 0  \\ 0 & 0 & 0 & 10^{ 2 } \end{pmatrix},
\end{equation*}

\begin{equation*}
    P = \begin{pmatrix} 5^{ 2 } & 0 & 0 & 0  \\ 0 & 10^{ 2 } & 0 & 0  \\ 0 & 0 & 5^{ 2 } & 0  \\0 & 0 & 0 & 10^{ 2 }  \end{pmatrix},
    R = \begin{pmatrix} 5^{ 2 } & 0  \\ 0 & 10^{ 2 } \end{pmatrix},
\end{equation*}

\begin{equation*}
     H = \begin{pmatrix}1 & 0 & 0 & 0  \\ 0 & 1 & 0 & 0 \end{pmatrix}.
\end{equation*}

We uniformly truncate the segmentation results under threshold values, which are $0.6$ for cars and $0.7$ for pedestrians.

%-------------------------------------------------------------------------
\begin{table*}[h]
\begin{center}
    \begin{tabular}{|cc|c|c|c|c|c|c|c|c|c|c|c|c|} 
     
        \hline 
       \multicolumn{2}{|c|}{\multirow{3}{*}{\footnotesize{\textbf{Trackers}}}} & 
       \multicolumn{4}{c|}{\footnotesize{\textbf{Modules}}} &  
       \multicolumn{8}{c|}{\footnotesize{\textbf{KITTI-MOTS Training Set}}} 
        \\ \cline{3-14}
         && \footnotesize{S2TA} & \multicolumn{2}{c|}{\footnotesize{Mask Merging}} & \footnotesize{T2TA} & 
        \multicolumn{4}{c|}{\footnotesize{Cars}} & \multicolumn{4}{c|}{\footnotesize{Pedestrians}}
         \\ \cline{3-14}
        && \footnotesize{MAF} & \footnotesize{IoU} & \footnotesize{Mask IoU} & \footnotesize{MAF} & 
        \footnotesize{sMOTSA$\uparrow$}&\footnotesize{MOTSA$\uparrow$} &  
        \footnotesize{IDS$\downarrow$}&  \footnotesize{FM$\downarrow$}& 
        \footnotesize{sMOTSA$\uparrow$}&\footnotesize{MOTSA$\uparrow$} &  
        \footnotesize{IDS$\downarrow$}&  \footnotesize{FM$\downarrow$}
        \\ \hline \hline
        \multicolumn{1}{|c|}{\multirow{6}{*}{\footnotesize{Ours}}} 
        & \footnotesize{$p1$}  
        &&&&& 
        \footnotesize{73.7} & \footnotesize{84.0}& \footnotesize{1322} & \footnotesize{1250} 
        & \footnotesize{56.4} & \footnotesize{71.2}& \footnotesize{800} & \footnotesize{721} 
        \\ \cline{2-14}
        & \multicolumn{1}{|c|}{\footnotesize{$p2$}}  
        &\footnotesize{\checkmark}&&&& 
        \footnotesize{76.3} & \footnotesize{86.6}& \footnotesize{642} & \footnotesize{606} 
        & \footnotesize{59.6} & \footnotesize{74.5}& \footnotesize{428} & \footnotesize{387} 
        \\ \cline{2-14}
         & \multicolumn{1}{|c|}{\footnotesize{$p3$}}  
        &\footnotesize{\checkmark}&\footnotesize{\checkmark}&&& 
        \footnotesize{76.8} & \footnotesize{86.5}& \footnotesize{598} & \footnotesize{572} 
        & \footnotesize{59.5} & \footnotesize{74.3}& \footnotesize{429} & \footnotesize{391} 
        \\ \cline{2-14}
         & \multicolumn{1}{|c|}{\footnotesize{$p4$}}  
        &\footnotesize{\checkmark}&&\footnotesize{\checkmark}&&
        \footnotesize{77.0} & \footnotesize{86.7}& \footnotesize{581} & \footnotesize{557} 
        & \footnotesize{59.6} & \footnotesize{74.4}& \footnotesize{423} & \footnotesize{382} 
        \\ \cline{2-14} 
         & \multicolumn{1}{|c|}{\footnotesize{$p5$}}  
        &\footnotesize{\checkmark}&&\footnotesize{\checkmark}&\
        \footnotesize{\checkmark}& 
        \footnotesize{77.8} & \footnotesize{87.6}& \footnotesize{362} & \footnotesize{518} 
        & \footnotesize{61.2} & \footnotesize{76.0}& \footnotesize{245} & \footnotesize{341} 
        \\ \cline{2-14} 
         & \multicolumn{1}{|c|}{\footnotesize{$p6$}}  
        &\footnotesize{\checkmark}&&\footnotesize{\checkmark}&\
        \footnotesize{\checkmark}& 
        \footnotesize{\textbf{78.5}} & \footnotesize{\textbf{88.1}}& \footnotesize{\textbf{212}} & \footnotesize{\textbf{419}} 
        & \footnotesize{\textbf{62.3}} & \footnotesize{\textbf{77.0}}& \footnotesize{\textbf{140}} & \footnotesize{\textbf{299}} 
        \\ \hline
    \end{tabular}
\end{center}
\caption{\textbf{Evaluation results on the KITTI-MOTS training set}. $p1$ is the baseline method without any proposed modules. In $p3$, $p4$, $p5$, and $p6$, the merging threshold $t_m$ is set to $0.4$. $p3$, $p4$, and $p5$ compute the appearance affinities of the bounding box regions, but $p6$ considers pixelwise mask-based foreground by setting the background pixels to zeros).}
\label{table:eval_train_kittimots}
\end{table*}
%-------------------------------------------------------------------------
\begin{table}[h]
\begin{center}
    \begin{tabular}{|cc|c|c|c|c|} 
     
        \hline 
       \multicolumn{2}{|c|}{\multirow{3}{*}{\footnotesize{\textbf{Trackers}}}} & 
       \multicolumn{4}{c|}{\footnotesize{\textbf{MOTSChallenge Training Set}}} 
        \\ \cline{3-6}
         && \multicolumn{4}{c|}{\footnotesize{Pedestrians}}
         \\ \cline{3-6}
        && \footnotesize{sMOTSA$\uparrow$}&\footnotesize{MOTSA$\uparrow$} &  
        \footnotesize{IDS$\downarrow$}&  \footnotesize{FM$\downarrow$}
        \\ \hline \hline
        \multicolumn{1}{|c|}{\multirow{5}{*}{\footnotesize{Ours}}} 
        & \footnotesize{$p1$}  
        & \footnotesize{64.5} & \footnotesize{75.9}& \footnotesize{686} & \footnotesize{604} 
        \\ \cline{2-6}
        & \multicolumn{1}{|c|}{\footnotesize{$p2$}}  
        &\footnotesize{64.5} & \footnotesize{75.9}& \footnotesize{535} & \footnotesize{487} 
        \\ \cline{2-6}
         & \multicolumn{1}{|c|}{\footnotesize{$p3$}}  
        & \footnotesize{64.6} & \footnotesize{75.9}& \footnotesize{565} & \footnotesize{523} 
        \\ \cline{2-6}
         & \multicolumn{1}{|c|}{\footnotesize{$p4$}}  
        & \footnotesize{65.0} & \footnotesize{76.3}& \footnotesize{539} & \footnotesize{497} 
        \\ \cline{2-6} 
         & \multicolumn{1}{|c|}{\footnotesize{$p5$}}  
        &\footnotesize{65.6} & \footnotesize{\textbf{77.1}}& \footnotesize{335} & \footnotesize{509} 
         \\ \cline{2-6} 
         & \multicolumn{1}{|c|}{\footnotesize{\boldmath{$p6$}}}  
        &\footnotesize{\textbf{65.8}} & \footnotesize{\textbf{77.1}}& \footnotesize{\textbf{262}} & \footnotesize{\textbf{465}} 
         
        \\ \hline
    \end{tabular}
\end{center}
\caption{\textbf{Evaluation results on the MOTSChallenge training set}. $p1$ is the baseline method without any proposed modules. In $p3$, $p4$, $p5$, and $p6$, the merging threshold $t_m$ is set to $0.4$. 
$p3$, $p4$, and $p5$ compute the appearance affinities of the bounding box regions but $p6$ considers the pixelwise mask-based foreground by setting the background pixels to zeros).}
\label{table:eval_train_motschallenge}
\end{table}

\begin{table}[h]
\caption{Threshold Settings for ``Mask Merging" \\ and ``Mask-Based Affinity Fusion (MAF)".}
\label{table:params}
\setlength{\tabcolsep}{3pt}
\begin{tabular}{|p{25pt}|p{142pt}|p{63pt}|}
\hline
\textbf{Symbol} & 
\textbf{Description} & 
\textbf{Value} \\
\hline
$\beta $& 
framewise motion update ratio in S2TA &
$\text{car: }0.4\text{, ped: }0.5$ \\
\hline
$t_{m} $& 
upper threshold for mask merging &
$\text{car: }0.3\text{, ped: }0.4$ \\
\hline
$f_{pm}$& 
upper threshold for $A_{pm}$ before MAF & 
$\text{car \& ped: }10^{-39}$ \\
\hline
$f_{appr}$&
upper threshold for $A_{appr}$ in MAF  & 
$\text{car \& ped: }0.85$ \\
\hline
%\multicolumn{3}{p{251pt}}{Vertical lines are optional in tables. Statements that serve as captions for 
%the entire table do not need footnote letters. }\\
\end{tabular}
\label{tab1}
\end{table}

\subsection{Analysis of Affinity Data} 
\label{subsec:anl_aff}
Figures~\ref{fig6}(a)-(c) and~\ref{fig7}(a)-(c) show that the position and motion affinity $A_{gmphd}$ and appearance affinity $A_{kcf}$ have quite different data magnitudes and distributions. In our experiments, $A_{pm}=\{10^{-9},\dots,10^{-3}\}$ and $A_{appr}=\{0.4,\dots,1.0\}$ are observed. 
Additionally, Figure~\ref{fig6}(a) shows that the cars have more concentrated distributions, with mean $m_{kcf}\approx0.944$ for appearance affinity than the pedestrians, with $m_{kcf}\approx0.905$ in Figure~\ref{fig7}(a). On the other hand, for the GMPHD affinity, pedestrians have more concentrated distributions as seen in Figures~\ref{fig6}(b) and Figure~\ref{fig7}(b). These facts are interpreted as follows: cars can be well discriminated by position and motion while pedestrians can be well discriminated by appearance. 

To considering these two characteristics, we propose MAF; from the distributions of normalized affinities $\bar{A}_{gmphd}$ and $\bar{A}_{kcf}$ in Figures~\ref{fig6}(d) and~\ref{fig7}(d), the gaps are much closer than before, and the two affinities are fused into $A_{maf}$ by using MAF.

\subsection{Ablation Studies}
\label{sub:ablation}
For the ablation studies, GMPHD\_MAF is evaluated on the training sequences of KITTI-MOTS and MOTSChallenge.
\noindent{\textbf{Key modules:}}
As discussed in Section~\ref{sec:proposed}, our method includes three key modules: HDA, mask merging, and MAF. HDA consists of S2TA and T2TA in order. Then, we can rearrange these modules with ``MAF in S2TA", ``Mask Merging", and ``MAF in T2TA" considering serial processes as described in Table~\ref{table:eval_train_kittimots}. Additionally, we can select either IoU~\eqref{eq:iou} or Mask IoU~\eqref{eq:mask_iou} for ``Mask Merging".

%Especially, ID-switch of the final model $p5$ decreases from 800 to 245 in KITTI-MOTS and from 686 to 335 in MOTSChallenge compare to $p1$. 
\noindent{\textbf{Experimental Studies on key parameters:}} We address experimental studies on key parameters in Appendix~\ref{apdx:params}, and the parameter settings are summarized in Table~\ref{table:params}. 

\noindent{\textbf{Effectiveness of the key modules:}} As seen in Tables~\ref{table:eval_train_kittimots} and~\ref{table:eval_train_motschallenge},
when the key modules ``MAF in S2TA", ``Mask Merging", and ``MAF in T2TA" are added to the baseline method $p1$ one by one, our method shows incremental and remarkable improvements. Comparing $p1$ and $p2$, $p1$ exploits one-step GMPHD filtering in computing only position and motion affinity, but $p2$ considers the position-motion affinity with the GMPHD filter and appearance affinity by the KCF in ``MAF in S2TA". The remarkable improvements in IDS and FM indicate that the proposed affinity fusion method works effectively. Comparing $p2$ and $p3$ in both Tables, because the results are advanced only in KITTI-MOTS, ``Mask Merging" may merge more than two segments of one object into one segment or not. However, we can see that at least Mask IoU works better than IoU in the merging of the results of $p3$ and $p4$. In $p5$ and $p6$, ``MAF in T2TA" is applied to our method, where $p5$ computes the appearance affinities for bounding box pixels, but $p6$ considers pixelwise mask-based foreground by setting the background pixels to zeros. Comparing the settings without T2TA, from $p2$ to $p4$, and with T2TA, $p5$ and $p6$, the results of $p5$ and $p6$ show that HDA with MAF reduces IDS very effectively in both datasets and mask-based affinity fusion works better than bounding box-based affinity fusion.

In summary, when adding the key modules ``MAF in S2TA", ``Mask Merging", and ``MAF in T2TA" one by one,
as shown in Tables~\ref{table:eval_train_kittimots} and~\ref{table:eval_train_motschallenge}, and Figure~\ref{fig11:vis_res}, our MOTS method shows incremental improvements from $p1$ to $p6$. The baseline method $p1$ is numerically improved as follows: for the KITTI-MOTS Cars training set, sMOTSA changes from 73.7 to 78.5 and IDS changes from 1,322 to 212; for the KITTI-MOTS Pedestrians training set, sMOTSA changes from 56.4 to 62.3 and IDS changes from 800 to 140; and for the MOTSChallenge training set, sMOTSA changes from 64.5 to 65.8 and IDS changes from 686 to 262.
Thus, we select $p6$ as the final model for testing and compare it with state-of-the-art methods. 

\subsection{Test Results}
\label{sub:test}
%-------------------------------------------------------------------------
\begin{table*}[h]
\begin{center}
    \begin{tabular}{|c|c|c|c|c|c|c|c|c|} 
        \hline 
       \multicolumn{1}{|c|}{\multirow{3}{*}{\footnotesize{\makecell{\textbf{Trackers}\\(online methods\\in \textbf{bold})}}}} & 
       \multicolumn{1}{c|}{\multirow{3}{*}{\footnotesize{\textbf{Det.}}}} & 
       \multicolumn{1}{c|}{\multirow{3}{*}{\footnotesize{\textbf{Seg.}}}} & 
       \multicolumn{6}{c|}{\footnotesize{\textbf{KITTI-MOTS Validation Set}}}
        \\ \cline{4-9}
         & & & \multicolumn{3}{c|}{\footnotesize{Cars}} & \multicolumn{3}{c|}{\footnotesize{Pedestrians}}
         \\ \cline{4-9}
         & &
        & \scriptsize{sMOTSA$\uparrow$} & \scriptsize{MOTSP$\uparrow$} & \scriptsize{IDS$\downarrow$} 
        & \scriptsize{sMOTSA$\uparrow$} & \scriptsize{MOTSP$\uparrow$} & \scriptsize{IDS$\downarrow$}
        \\ \hline \hline
     
        \footnotesize{TrackRCNN~\cite{mots_trackrcnn}} 
        & TrackRCNN & TrackRCNN
        &  \footnotesize{76.2} & \footnotesize{87.2}& \footnotesize{83} 
        & \footnotesize{47.1} & \footnotesize{75.7}& \footnotesize{78} 
        \\ \hline
        
        \footnotesize{CAMOT~\cite{camot}} 
        & TrackRCNN & TrackRCNN
        &  \footnotesize{67.4} & \footnotesize{86.5}& \footnotesize{220} 
        & \footnotesize{39.5} & \footnotesize{73.1}& \footnotesize{131} 
        \\ \hline
        
        \footnotesize{CIWT~\cite{ciwt}} 
        & TrackRCNN & TrackRCNN
        &  \footnotesize{68.1} & \footnotesize{76.8}& \footnotesize{\textcolor{red}{10}} 
        & \footnotesize{42.9} & \footnotesize{75.7}& \footnotesize{42} 
        \\ \hline
        
        \footnotesize{\textbf{BeyondPixels~\cite{beyondpixels}}} 
        & RRC~\cite{rrc} & TrackRCNN
        &  \footnotesize{76.9} & \footnotesize{86.5}& \footnotesize{88} 
        & \footnotesize{-} & \footnotesize{-}& \footnotesize{-} 
        \\ \hline
        
        \footnotesize{MOTSNet~\cite{motsnet}} 
        & MOTSNet & MOTSNet
        &  \footnotesize{78.1} & \footnotesize{86.9}& \footnotesize{-} 
        & \footnotesize{54.6} & \footnotesize{\textcolor{blue}{79.7}}& \footnotesize{-} 
        \\ \hline
        
         \multirow{1}{*}{\footnotesize{\textbf{MOTSFusion~\cite{MOTSFusionJ}}}}
        & TrackRCNN & TrackRCNN
        &  \footnotesize{78.2} & \footnotesize{-}& \footnotesize{36} 
        & \footnotesize{50.1} & \footnotesize{-}& \footnotesize{\textcolor{blue}{34}} 
        \\ \hline %\cline{2-9}
        
        %& RRC & BB2SegNet~\cite{bb2segnet}
       %&  \footnotesize{\textcolor{red}{85.7}} & \footnotesize{-}& \footnotesize{31} 
       % & \footnotesize{-} & \footnotesize{-}& \footnotesize{-}
       %\\ \hline
        
        \footnotesize{\textbf{PointTrack~\cite{PointTrack}}} 
        & PointTrack & PointTrack 
        & \footnotesize{\textcolor{red}{85.5}} & \footnotesize{-}& \footnotesize{\textcolor{blue}{22}}
        & \footnotesize{\textcolor{red}{62.4}} & \footnotesize{-}& \footnotesize{\textcolor{red}{19}} 
        \\ \hline
         
        \multirow{4}{*}{\footnotesize{\textbf{GMPHD\_MAF ($\textbf{\textit{p6}}$)}} }
        & TrackRCNN & TrackRCNN
        &  \footnotesize{76.9} & \footnotesize{87.1}& \footnotesize{68}
        & \footnotesize{48.8} & \footnotesize{76.4}&
        \footnotesize{41} 
        \\ \cline{2-9}
        & MaskRCNN~\cite{maskrcnn} & MaskRCNN
        &  \footnotesize{82.0} & \footnotesize{\textcolor{blue}{90.3}}& \footnotesize{59}
        & \footnotesize{\textcolor{blue}{59.9}} & \footnotesize{\textcolor{red}{82.8}}& \footnotesize{49} 
        \\ \cline{2-9}
        & RRC & BB2SegNet~\cite{bb2segnet}
        &  \footnotesize{{84.5}} & \footnotesize{\textcolor{red}{90.8}}& \footnotesize{49}
        & \footnotesize{-} & \footnotesize{-}& \footnotesize{-}  
        \\ \cline{2-9}
        & PointTrack & PointTrack
        &  \footnotesize{\textcolor{blue}{85.0}} & \footnotesize{\textcolor{blue}{90.3}}& \footnotesize{66}
        & \footnotesize{-} & \footnotesize{-}& \footnotesize{-}
        \\ \hline
    \end{tabular}
\end{center}
\caption{\textbf{Evaluation results on the KITTI-MOTS validation set}. Our method is denoted as GMPHD\_MAF.
\\The $1^{st}$ and $2^{nd}$ best scores are highlighted in \textcolor{red}{red} and \textcolor{blue}{blue}, respectively. TrackRCNN~\cite{mots_trackrcnn} and MaskRCNN~\cite{maskrcnn} are the public detection \& segmentation methods but the others are privately refined segmentation results.}
\label{table:eval_valid_kitti}
\end{table*}
%-------------------------------------------------------------------------
\begin{table*}[h]
\begin{center}
    \begin{tabular}{|c|c|c|c|c|c|c|c|c|c|c|c|c|} 
        \hline 
       \multicolumn{1}{|c|}{\multirow{3}{*}{\footnotesize{\makecell{\textbf{Trackers}\\(online methods\\in \textbf{bold})}}}} & 
       \multicolumn{8}{c|}{\footnotesize{\textbf{KITTI-MOTS Test Set}}} &
       \multicolumn{4}{c|}{\footnotesize{\textbf{MOTSChallenge Test Set}}} 
        \\ \cline{2-13}
        & \multicolumn{4}{c|}{\footnotesize{Cars}} & \multicolumn{4}{c|}{\footnotesize{Pedestrians}}
         & \multicolumn{4}{c|}{\footnotesize{Pedestrians}}  
         \\ \cline{2-13}
        & \scriptsize{sMOTSA$\uparrow$} & \scriptsize{MOTSP$\uparrow$} & 
        \scriptsize{IDS$\downarrow$} & \scriptsize{FPS$\uparrow$} 
        & \scriptsize{sMOTSA$\uparrow$} & \scriptsize{MOTSP$\uparrow$} & 
        \scriptsize{IDS$\downarrow$} & \scriptsize{FPS$\uparrow$}
        & \scriptsize{sMOTSA$\uparrow$} & \scriptsize{MOTSP$\uparrow$} & 
        \scriptsize{IDS$\downarrow$} & \scriptsize{FPS$\uparrow$}
        \\ \hline \hline
     
        \footnotesize{TrackRCNN~\cite{mots_trackrcnn}} 
        &  \footnotesize{67.0} & \footnotesize{85.1}& \footnotesize{692}& \footnotesize{2.0} 
        & \footnotesize{47.3} & \footnotesize{74.6}& \footnotesize{481}& \footnotesize{2.0} 
        & \footnotesize{40.6} & \footnotesize{76.1} & \footnotesize{576}& \footnotesize{\textcolor{blue}{2.0}} 
        \\ \hline
       \footnotesize{\textbf{MOTSFusion~\cite{MOTSFusionJ}}} 
       &  \footnotesize{75.0} & \footnotesize{\textcolor{red}{89.3}}& \footnotesize{\textcolor{blue}{201}}& \footnotesize{2.3}
        & \footnotesize{58.7} & \footnotesize{81.5}& \footnotesize{\textcolor{blue}{279}}& \footnotesize{2.3} 
        & \footnotesize{-} & \footnotesize{-} & \footnotesize{-}& \footnotesize{-}
        \\ \hline
        \footnotesize{ReMOTS~\cite{ReMOTS}} 
       &  \footnotesize{75.9} & \footnotesize{88.2}& \footnotesize{716}& \footnotesize{0.3}
        & \footnotesize{\textcolor{red}{66.0}} & \footnotesize{82.0}& \footnotesize{391}& \footnotesize{0.3} 
        & \footnotesize{\textcolor{red}{70.4}} & \footnotesize{\textcolor{blue}{84.0}} & \footnotesize{\textcolor{blue}{231}}& \footnotesize{0.3}  
        \\ \hline
        \footnotesize{\textbf{PointTrack~\cite{PointTrack}}} 
       &  \footnotesize{\textcolor{red}{78.5}} & \footnotesize{87.1}& \footnotesize{\textcolor{red}{114}}& \footnotesize{\textcolor{red}{22.2}}
        & \footnotesize{61.5} & \footnotesize{\textcolor{red}{82.4}}& \footnotesize{632}& \footnotesize{\textcolor{red}{22.2}} 
        & \footnotesize{58.0} & \footnotesize{-} & \footnotesize{-}& \footnotesize{-} 
        \\ \hline
         \multirow{1}{*}{\footnotesize{\textbf{GMPHD\_MAF ($\textbf{\textit{p6}}$)}} }
        &  \footnotesize{\textcolor{blue}{76.7}} & \footnotesize{\textcolor{blue}{88.4}}& \footnotesize{430}& \footnotesize{\textcolor{blue}{7.7}}
        & \footnotesize{\textcolor{blue}{65.2}} & \footnotesize{\textcolor{blue}{82.3}}& \footnotesize{\textcolor{red}{277}}& \footnotesize{\textcolor{blue}{19.4}} 
         & \footnotesize{\textcolor{blue}{69.4}} & \footnotesize{\textcolor{red}{84.2}} & \footnotesize{\textcolor{blue}{484}}& \footnotesize{\textcolor{red}{2.6}} 
        \\ \hline
    \end{tabular}
\end{center}
\caption{\textbf{Evaluation results on the KITTI-MOTS and MOTSChallenge test sets}. Ours is denoted by GMPHD\_MAF.
All methods except PointTrack~\cite{PointTrack} exploits instance segmentation results of MaskRCNN~\cite{maskrcnn} (public segmentation) as an input.
\\The $1^{st}$ and the $2^{nd}$ best scores are highlighted in \textcolor{red}{red} and \textcolor{blue}{blue}, respectively.}
\label{table:eval_test}
\end{table*}
%-------------------------------------------------------------------------
\begin{table}[h]
\begin{center}
    \begin{tabular}{|c|c|c|c|} 
        \hline 
       \multicolumn{1}{|c|}{\multirow{3}{*}{\footnotesize{\makecell{\textbf{Trackers}\\(online methods\\in \textbf{bold})}}}} & 
       \multicolumn{3}{c|}{\footnotesize{\textbf{MOTSChallenge Training Set}}}
        \\ \cline{2-4}
         & \multicolumn{3}{c|}{\footnotesize{Pedestrians}}
         \\ \cline{2-4}
        & \scriptsize{sMOTSA$\uparrow$} & \scriptsize{MOTSA$\uparrow$} & \scriptsize{MOTSP$\uparrow$} 
        \\ \hline \hline
     
        \footnotesize{MHT-DAM~\cite{mhtdam}} 
        &  \footnotesize{48.0} & \footnotesize{62.7}& \footnotesize{79.8}
        \\ \hline
        \footnotesize{FWT~\cite{fwt}} 
        &  \footnotesize{49.3} & \footnotesize{64.0}& \footnotesize{79.7}
        \\ \hline
        \footnotesize{\textbf{MOTDT~\cite{motdt}}} 
        &  \footnotesize{47.8} & \footnotesize{61.1}& \footnotesize{80.0}
        \\ \hline
        \footnotesize{jCC~\cite{jcc}} 
        &  \footnotesize{48.3} & \footnotesize{63.0}& \footnotesize{79.9}
        \\ \hline
         \footnotesize{TrackRCNN~\cite{mots_trackrcnn}} 
        &  \footnotesize{52.7} & \footnotesize{66.9}& \footnotesize{80.2}
        \\ \hline
       \footnotesize{MOTSNet~\cite{motsnet}} 
       &  \footnotesize{56.8} & \footnotesize{69.4}& \footnotesize{\textcolor{blue}{82.7}}
        \\ \hline
        \footnotesize{\textbf{PointTrack~\cite{PointTrack}}} 
       &  \footnotesize{\textcolor{blue}{58.1}} & \footnotesize{\textcolor{blue}{70.6}}& \footnotesize{-}
       \\ \hline
        %\footnotesize{\textbf{GMPHD\_MAF ($\textbf{\textit{p5}}$})} 
       % &  \footnotesize{65.6} & \footnotesize{\textbf{77.1}}& \footnotesize{86.1}
       % \\ \hline
        \footnotesize{\textbf{GMPHD\_MAF ($\textbf{\textit{p6}}$})} 
        &  \footnotesize{\textcolor{red}{65.8}} & \footnotesize{\textcolor{red}{77.1}}& \footnotesize{\textcolor{red}{86.1}}
        \\ \hline
    \end{tabular}
\end{center}
\caption{\textbf{Evaluation results on  MOTSChallenge training (=validation) set}. Ours is denoted by GMPHD\_MAF. All methods except \cite{PointTrack,motsnet} exploit the instance segmentation results of MaskRCNN~\cite{maskrcnn} (public segmentation) as an input.
\\The $1^{st}$ and $2^{nd}$ best scores are highlighted in \textcolor{red}{red} and \textcolor{blue}{blue}, respectively.}
\label{table:eval_valid_mots}
\end{table}

We evaluate our proposed method compared to state-of-the-art MOTS methods~\cite{MOTSFusionJ,mots_trackrcnn,ReMOTS,PointTrack} in the validation and test sets of the KITTI-MOTS and MOTSChallenge benchmarks. Each dataset has a training set, including a validation set and a test set as addressed in Table~\ref{table:dataset}. The reason we also use the validation sets for evaluation is that some MOTS methods~\cite{camot,ciwt,beyondpixels} provide only the evaluation results in validation sets. In addition, state-of-the-art 2D MOT methods~\cite{mhtdam, fwt, motdt, jcc} are compared to ours by matching their 2D MOT results in bounding boxes with the 2D pixelwise masks of TrackRCNN~\cite{mots_trackrcnn}. Tables~\ref{table:eval_valid_kitti},~\ref{table:eval_valid_mots}, and~\ref{table:eval_test} show the evaluation results. In the three tables, our method is denoted as GMPHD\_MAF, fully online methods are in~\textbf{bold}, and others, i.e., near-online or offline methods, are not. A fully online approach uses only past and present frames, near-online uses a small batch of future frames, and offline uses all frames at once.

\noindent{\textbf{KITTI-MOTS:}}
For comparison in KITTI-MOTS, we use not only the test set but also the validation set because many state-of-the-art methods, such as~\cite{camot,ciwt,beyondpixels,motsnet} only provide the evaluation results in the validation set. The specifications of KITTI-MOTS sequences are addressed in Table ~\ref{table:dataset}. Other methods~\cite{mots_trackrcnn,MOTSFusionJ,PointTrack,ReMOTS} also provide the evaluation results in the test set.
In the validation set, the developed method shows the second best sMOTSA and MOTSP scores for cars, and the second best sMOTSA and the best MOTSP for pedestrians among all methods but the best scores compared to the public detection and segmentation results: TrackRCNN- and MaskRCNN-based MOTS methods~\cite{mots_trackrcnn,camot,ciwt,MOTSFusionJ} for cars and pedestrians (see Table~\ref{table:eval_valid_kitti}). \cite{beyondpixels,motsnet,PointTrack} adapt refinement techniques to TrackRCNN or MaskRCNN. 
In the test set, our method shows that the second best sMOTSA and MOTSA scores among all methods for cars but the best scores against all online methods for pedestrians (see Table~\ref{table:eval_test}). Even if our method shows a lower sMOTSA score, $59.4$, than PointTrack~\cite{PointTrack}, $62.4$, for pedestrians in the validation, ours achieves a better sMOTSA, $65.2$, than PointTrack~\cite{PointTrack}, $61.5$, in the test. 

\noindent{\textbf{MOTSChallenge:}}
This dataset contains only pedestrians, and from the results of Tables~\ref{table:eval_valid_mots} and~\ref{table:eval_test}, we can see that the proposed method is more competitive for pedestrians against state-of-the-art methods~\cite{mhtdam,fwt,motdt,jcc,mots_trackrcnn,motsnet,MOTSFusionJ,PointTrack,ReMOTS} in the validation set. In the test sequences, our method achieves the second best sMOTSA, $69.4$, in MOTSChallenge, and the best is $70.4$, achieved by ReMOTS~\cite{ReMOTS}. However, ReMOTS shows a lower sMOTSA than ours for cars in KITTI-MOTS (see Table~\ref{table:eval_test}). This means that ours is comparable with offline methods.

In summary, referring to Tables~\ref{table:eval_valid_kitti},~\ref{table:eval_valid_mots}, and~\ref{table:eval_test}, our proposed method, named GMPHD\_MAF, not only shows the best sMOTSA scores among online methods using public segmentation without a refinement process but also achieves competitive performance against state-of-the-art online and offline MOTS methods~\cite{camot,ciwt,beyondpixels,mhtdam,fwt,motdt,jcc,mots_trackrcnn,motsnet,MOTSFusionJ,PointTrack,ReMOTS}. In particular, comparing our method to these methods, relatively better performances are measured for pedestrians than for cars. We infer that this is due to the difference in the rigidity of each object. We think cars have relatively rigid shapes the so tracking performance may be greatly affected by the accuracy of segmentation for position affinity, while pedestrians have nonrigid shapes so appearance affinity may affect the performance more than segmentation does. We discussed this in~\ref{subsec:anl_aff}.
In the KITTI-MOTS test set, GMPHD\_MAF ($p6$) runs at $7.7$ FPS for cars and $19.4$ FPS for pedestrians when operating separately. When tracking both cars and pedestrians, our method runs at $5.3$ FPS with serial processing but $6.3$ FPS with parallel processing, which is approximately $20$ percent faster than the former. In the MOTSChallenge test set, our method shows the highest speed even though it runs at $2.6$ FPS, since the test set consists of three high-resolution scenes ($1920$x$1080$) and one normal-resolution scene ($640$x$480$), as described in Table~\ref{table:dataset}.
%For KITTI-MOTS, we also implement and execute our method that can run in parallel processing for cars and pedestrians, simultaneously.

\section{Conclusions}
\label{sec:conclusion}
In this paper, we propose a highly practical MOTS method named GMPHD\_MAF, which is a feasible and easily reproducible combination of four key modules: a GMPHD filter, HDA, mask merging, and MAF. These key modules can operate in the proposed fully online MOTS framework which tracks cars and pedestrians in parallel CPU-only processes. 
These modules show incremental improvements in evaluation on the training sets of KITTI-MOTS and MOTSChallenge in terms of MOTS measures such as sMOTSA, MOTSA, IDS, FM, and FPS. In the validation and test sets of the two popular datasets, GMPHD\_MAF achieves very competitive performance against the state-of-the-art MOTS methods. In future work, we expect that the proposed MOTS method will be reproduced and extended with a more precise or simpler position and motion filtering model and more rapid or sophisticated appearance feature extractors such as deep neural network-based re-identification techniques.

% or


\appendices
\section{Dataset Specifications}
Table~\ref{table:dataset} describes the KITTI-MOTS and MOTSChallenge benchmark datasets in terms of training, validation (Valid), and test sequences, frames per second (FPS), resolution, and the number of frames (Frame). The ablation studies and experimental results using these datasets are presented in Tables~\ref{table:eval_train_kittimots},~\ref{table:eval_train_motschallenge},~\ref{table:eval_valid_kitti},~\ref{table:eval_test}, and ~\ref{table:eval_valid_mots} of Section~\ref{sec:experiments}.

\begin{table}[h]
\centering
\caption{Dataset specifications for KITTI-MOTS and MOTSChallenge.}
\label{table:dataset}
 \begin{tabular}{|c|c|c|c|c|c|} 
 \hline
 \multicolumn{1}{|c|}{Dataset} & \multicolumn{1}{c|}{Set} & \multicolumn{1}{c|}{Sequence} & \multicolumn{1}{c|}{FPS} & \multicolumn{1}{c|}{Resolution} & \multicolumn{1}{c|}{Frame}  \\ \hline\hline
 
 % KITTI-MOTS training and validation
\multirow{6}{*}{\rotatebox[origin=c]{90}{KITTI-MOTS}}

& \rotatebox[origin=c]{90}{\scriptsize{Train}}
& \scriptsize{\makecell{kitti-tracking-train:\\ 00$\sim$20}} & 10 &\scriptsize{\makecell{1224x370, \\1238x374, \\, or 1242x375}}
& 8,008   \\ \cline{2-6}

&\rotatebox[origin=c]{90}{\scriptsize{Valid}} 
& \scriptsize{\makecell{kitti-tracking-train:\\ 02, 06, 07, 08,\\10, 13, 14, 16, 18}} & 10 &\scriptsize{\makecell{1224x370, \\1238x374, \\, or 1242x375}}
& 2,981  \\ \cline{2-6}

%  KITTI -MOTS test
& \rotatebox[origin=c]{90}{Test}
& \scriptsize{\makecell{kitti-tracking-test:\\ 00$\sim$28}} & 10 & \scriptsize{\makecell{1224x370, \\1238x374, \\, or 1242x375}}
& 11,095 \\ \hline\hline

% MOTSChallenge training (=validation)
\multirow{10}{*}{\rotatebox[origin=c]{90}{MOTSChallenge}}
& \multirow{4}{*}{\rotatebox[origin=c]{90}{\scriptsize{Train}}}
& \scriptsize{MOTS20-02} & 30 &\scriptsize{1920x1080}& 600 \\ \cline{3-6}
&& \scriptsize{MOTS20-05} & 14 &\scriptsize{640x480}& 837 \\ \cline{3-6}
&& \scriptsize{MOTS20-09} & 30 &\scriptsize{1920x1080}& 525 \\ \cline{3-6}
&& \scriptsize{MOTS20-11} & 30 &\scriptsize{1920x1080}& 900 \\ \cline{2-6}
& \multicolumn{4}{c|}{Total Frames} & 2,862 \\ \cline{2-6}

% MOTSChallenge-test
&  \multirow{4}{*}{\rotatebox[origin=c]{90}{\scriptsize{Test}}}
& \scriptsize{MOTS20-01} & 30&\scriptsize{1920x1080}& 450 \\ \cline{3-6}
&& \scriptsize{MOTS20-06} & 14 &\scriptsize{640x480}& 1,194  \\\cline{3-6}
&& \scriptsize{MOTS20-07} & 30 &\scriptsize{1920x1080}& 500  \\ \cline{3-6}
&& \scriptsize{MOTS20-12} & 30 &\scriptsize{1920x1080}& 900  \\ \cline{2-6}

& \multicolumn{4}{c|}{Total Frames} & 3,044 \\ \hline

\end{tabular}
\end{table}

\section{Benchmark Leaderboards}
The full benchmark results, including published and unpublished methods are available at the online leaderboards below.

\footnotesize{\textbf{KITTI-MOTS}:~\href{http://www.cvlibs.net/datasets/kitti/old\_eval\_mots.php}{\underline{http://www.cvlibs.net/datasets/kitti/old\_eval\_mots.php}}}

\footnotesize{\textbf{MOTSChallenge}:~\href{https://motchallenge.net/results/MOTS/}{\underline{https://motchallenge.net/results/MOTS/}}}
\section{Experimental Studies on the Parameters}
\label{apdx:params}

\begin{figure}[h]
\begin{center}
   \includegraphics[width=8.5cm]{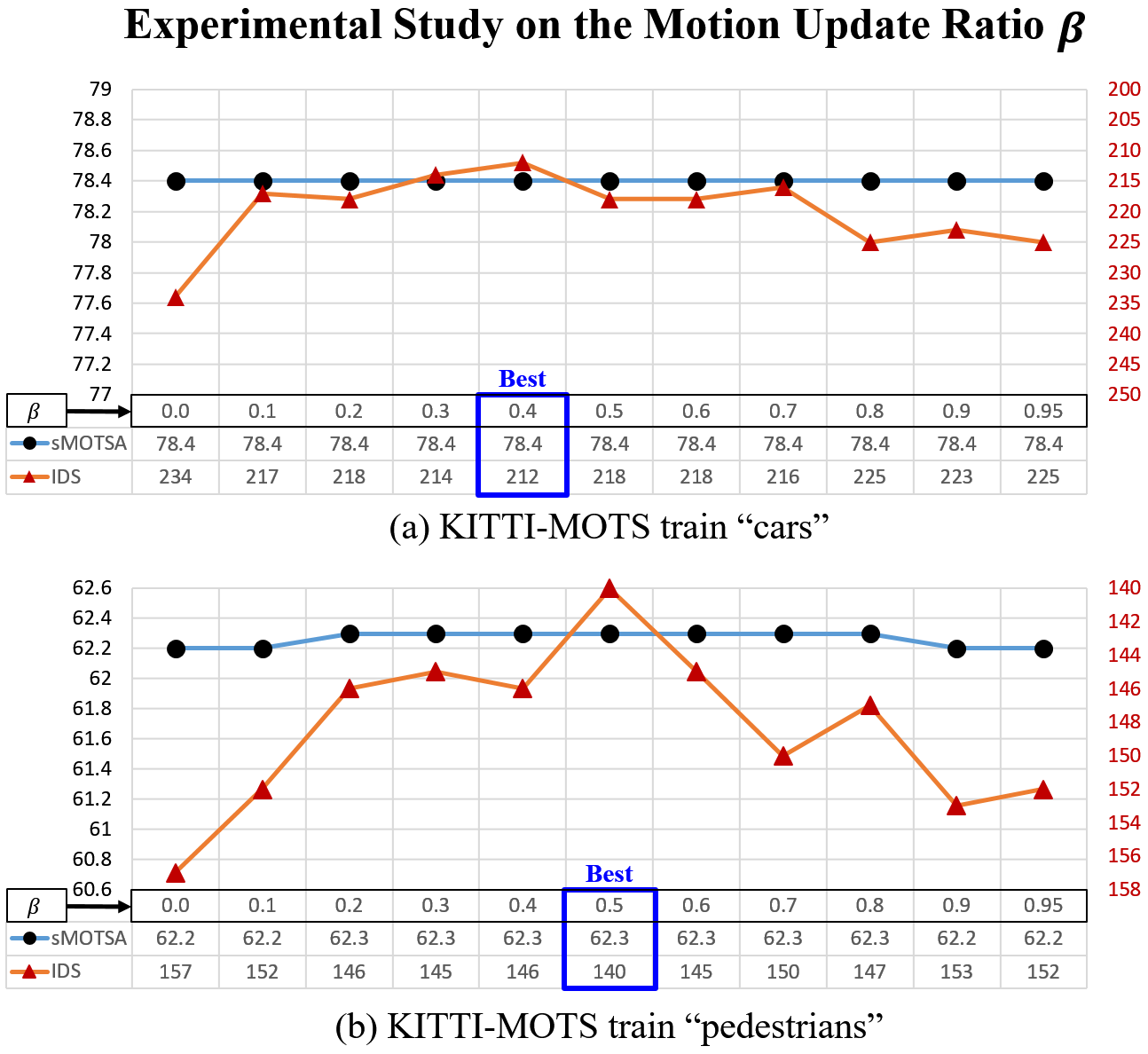}
\end{center}
   \caption{The framewise motion update ratio $\beta$ is presented in~\eqref{eq25}. In the above experiments on the KITTI-MOTS training set, the best sMOTSA and IDS scores are achieved when $\beta$ is $0.4$ and $0.5$ for cars and pedestrians, respectively. The same values are set for the test, and the results are presented in Tables~\ref{table:eval_valid_kitti} and~\ref{table:eval_test}.}
\label{fig8}
\end{figure}

\begin{figure}[h]
\begin{center}
   \includegraphics[width=8.5cm]{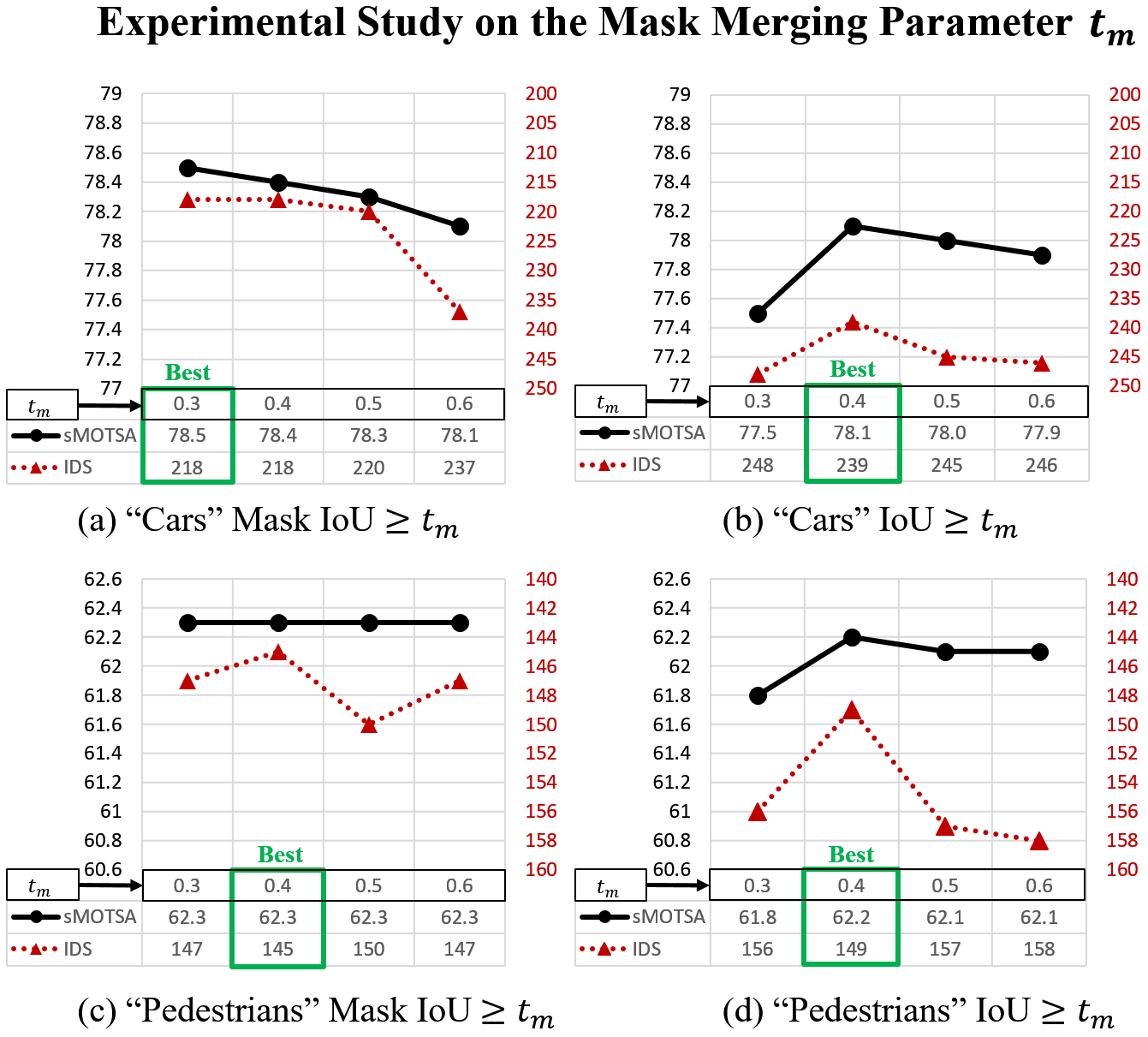}
\end{center}
   \caption{The mask merging threshold $t_m$ is presented in Subsection~\ref{subsec:merging}. In the above experiments on the KITTI-MOTS training set, using mask IoU (a)(c), the best sMOTSA and IDS scores are achieved when $\beta$ is $0.3$ and $0.4$ for cars and pedestrians, respectively. Using IoU (b)(d), the best sMOTSA and IDS scores are achieved when $\beta$ is $0.4$ and $0.4$ for cars and pedestrians, respectively. Comparing (a)(c) and (b)(d), mask IoU shows not only less sensitivity to the parameter values but also better performance than IoU. Thus, mask IoU is selected for the mask merging measure, the same values are set for the test and the results are presented in Tables~\ref{table:eval_valid_kitti} and~\ref{table:eval_test}}
\label{fig9}
\end{figure}

\begin{figure}[h]
\begin{center}
   \includegraphics[width=8.5cm]{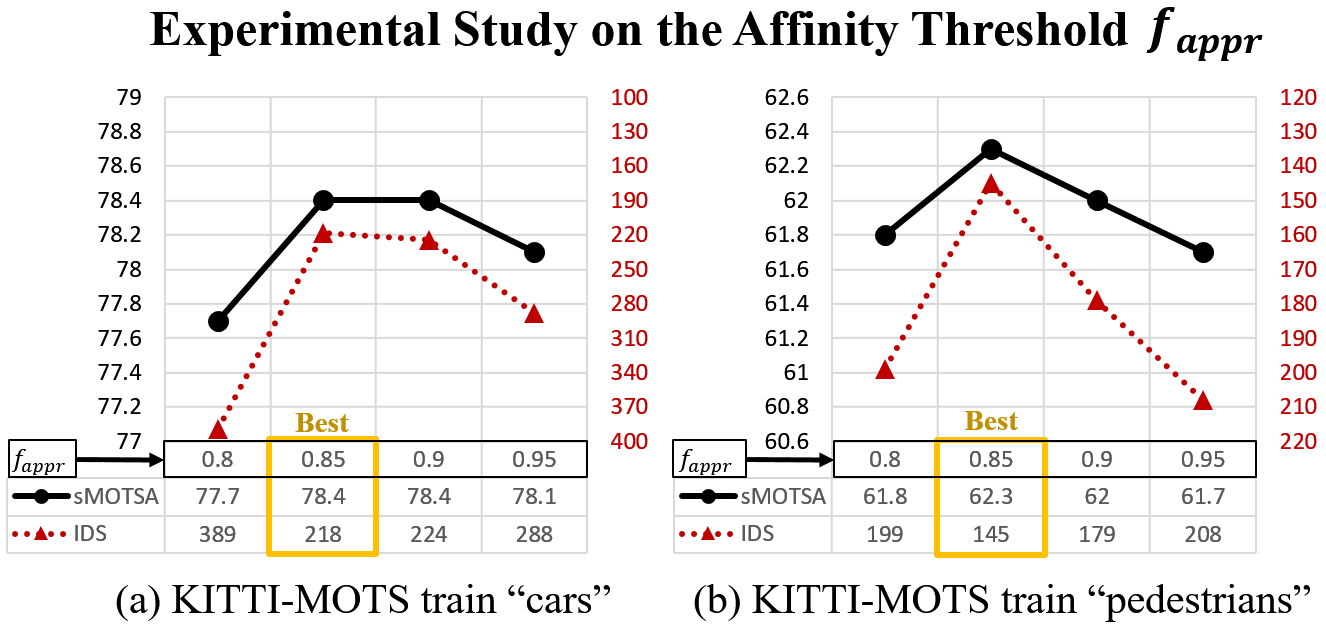}
\end{center}
   \caption{The appearance affinity threshold $f_{appr}$ is used for high $A_{appr}$ value filtering before MAF (see~\eqref{eq:maf}). When $A_{appr}$~\eqref{eq:affinityappr} between a state and an observation is greater than or equal to $f_{appr}$, the pair is considered for association even if $A_{pm}$ is less than $f_{pm}$ (see Table~\ref{table:params}). In the above experiments on the KITTI-MOTS training set, the best sMOTSA and IDS scores are achieved when $f_{appr}$ is $0.85$ for both cars and pedestrians. The same values are set for the test, and the results are presented in Tables~\ref{table:eval_valid_kitti} and~\ref{table:eval_test}.}
\label{fig10}
\end{figure}

\begin{figure*}[h]
\begin{center}
   \includegraphics[width=18cm]{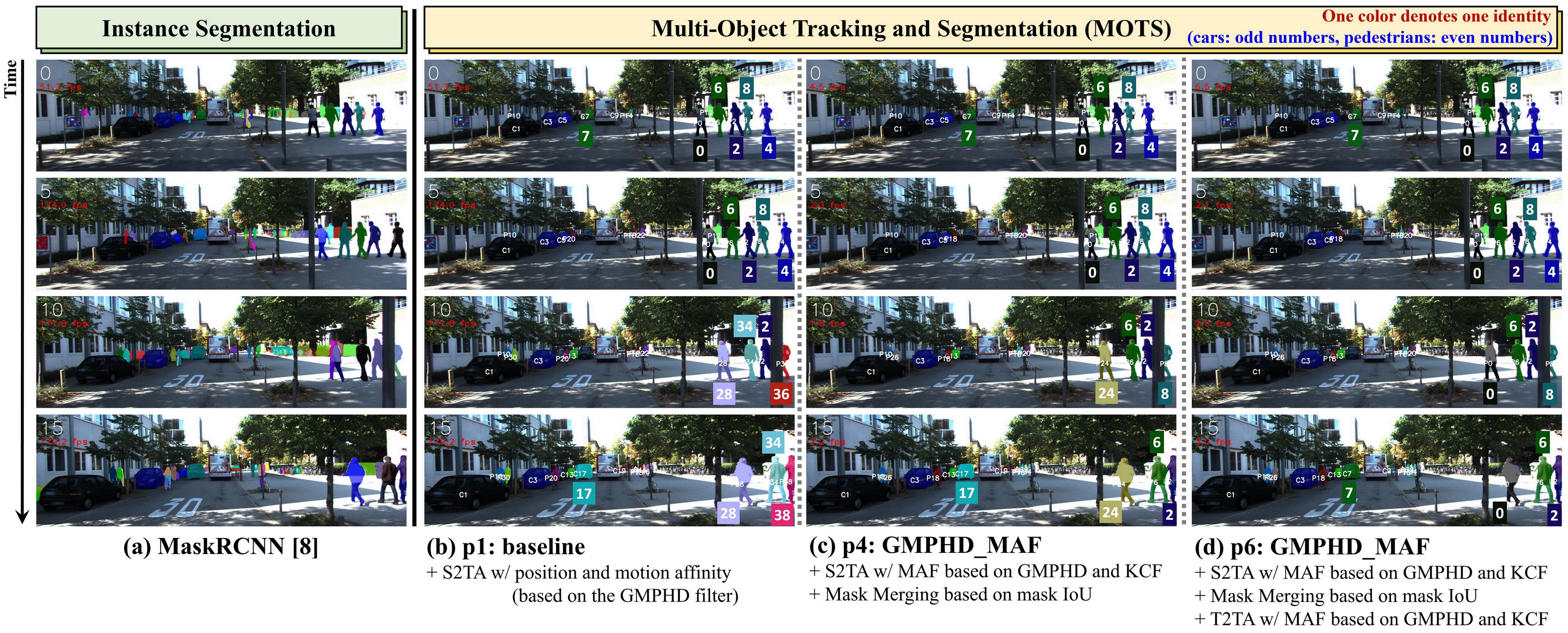}
\end{center}
\caption{Visualization of the segmentation and MOTS results on KITTI-MOTS test sequence 0018. (b), (c), and (d) are the results of the three different settings of the proposed method, which are based on the same segmentation results (a) from MaskRCNN~\cite{maskrcnn}. Comparing (b) the baseline model $p1$ and (c) the model $p4$ with S2TA and mask merging (without T2TA), in (b), the IDs, ``0, 2, 4, 6, 8", of the five pedestrians at the right side of the scene are switched except the person w/ ID 2, but, in (c), only the pedestrian w/ ID 0 gets switched to ID 24. In (d) the final model $p6$, the five IDs are preserved since T2TA can find the IDs after occlusion with trees at the right side. In addition, the car w/ ID 7 at frame 0 are recovered at frame 15, while (b) and (c) do not recover the car w/ ID 7 that are switched to ID 17.}
\label{fig11:vis_res}
\end{figure*}

\normalsize{
Experimental studies of key parameters of our method are presented in Figure~\ref{fig8}, ~\ref{fig9}, and~\ref{fig10}. The parameters are addressed in Section~\ref{sec:proposed} and summarized in Table~\ref{table:params}.
}

%\begin{figure}[h]
%\begin{center}
%   \includegraphics[width=8.5cm]{figures/fig12.jpg}
%\end{center}
%   \caption{The appearance affinity threshold $f_{appr}$ is used for high $A_{appr}$ value filtering before MAF (see~\eqref{eq:maf}). When $A_{appr}$~\eqref{eq:affinityappr} between a state and an observation is greater than or equal to $f_{appr}$, the pair is considered for association even if $A_{pm}$ is less than $f_{pm}$ (see Table~\ref{table:params}). In the above experiments on the MOTSChallenge training set, the best sMOTSA and IDS scores are achieved when $f_{appr}$ is $0.5$ for pedestrians. The same values are set for the test, and the results are presented in Tables~\ref{table:eval_valid_mots} and~\ref{table:eval_test}.}
%\label{fig11}
%\end{figure}

\bibliographystyle{IEEEtran}
\bibliography{draft_arXiv}

\begin{thebibliography}{10}
\providecommand{\url}[1]{#1}
\csname url@samestyle\endcsname
\providecommand{\newblock}{\relax}
\providecommand{\bibinfo}[2]{#2}
\providecommand{\BIBentrySTDinterwordspacing}{\spaceskip=0pt\relax}
\providecommand{\BIBentryALTinterwordstretchfactor}{4}
\providecommand{\BIBentryALTinterwordspacing}{\spaceskip=\fontdimen2\font plus
\BIBentryALTinterwordstretchfactor\fontdimen3\font minus
  \fontdimen4\font\relax}
\providecommand{\BIBforeignlanguage}[2]{{%
\expandafter\ifx\csname l@#1\endcsname\relax
\typeout{** WARNING: IEEEtran.bst: No hyphenation pattern has been}%
\typeout{** loaded for the language `#1'. Using the pattern for}%
\typeout{** the default language instead.}%
\else
\language=\csname l@#1\endcsname
\fi
#2}}
\providecommand{\BIBdecl}{\relax}
\BIBdecl

\bibitem{mot15}
A.~Milan, L.~Leal-Taix\'e, I.~Reid, S.~Roth, and K.~Schindler, ``{MOTChallenge}
  2015: Towards a benchmark for multi-target tracking,'' 2015, [{O}nline].
  Available: ar{X}iv:1504.01942.

\bibitem{kitti}
A.~Geiger, P.~Lenz, and R.~Urtasun, ``Are we ready for autonomous driving? the
  {KITTI} vision benchmark suite,'' in \emph{Proc. IEEE Conf. Comput. Vis.
  Pattern Recognit. (CVPR)}, Jun. 2012, pp. 3354--3361.

\bibitem{mot16}
L.~Leal-Taix\'e, A.~Milan, I.~Reid, S.~Roth, and K.~Schindler, ``{MOT16}: A
  benchmark for multi-object tracking,'' 2016, [{O}nline]. Available:
  ar{X}iv:1603.00831.

\bibitem{pointpillars}
A.~H. Lang, S.~Vora, H.~Caesar, L.~Zhou, J.~Yang, and O.~Beijbom,
  ``Point{P}illars: Fast encoders for object detection from point clouds,'' in
  \emph{Proc. IEEE Conf. Comput. Vis. Pattern Recognit. (CVPR)}, Jun. 2019, pp.
  12\,697--12\,705.

\bibitem{yolo}
J.~Redemon, S.~Divvala, R.~Girshick, and A.~Farhadi, ``You only look once:
  Unified, real-time object detection,'' in \emph{Proc. IEEE Conf. Comput. Vis.
  Pattern Recognit. (CVPR)}, Jul. 2016, pp. 779--788.

\bibitem{frcnn}
S.~Ren, K.~He, R.~Girshick, , and J.~Sun, ``Faster {R}-{CNN}: Towards real-time
  object detection with region proposal networks,'' in \emph{Proc. 28th Int.
  Conf. Neural Inf. Process. Syst. (NIPS)}, Dec. 2015, pp. 91--99.

\bibitem{pointrcnn}
S.~Shi, X.~Wang, and H.~Li, ``Point{RCNN}: 3{D} object proposal generation and
  detection from point cloud,'' in \emph{Proc. IEEE Conf. Comput. Vis. Pattern
  Recognit. (CVPR)}, Jun. 2019, pp. 770--779.

\bibitem{maskrcnn}
K.~He, G.~Gkioxari, P.~Doll\'ar, and R.~Girshick, ``Mask {R}-{CNN},'' in
  \emph{Proc. IEEE Int. Conf. Comput. Vis. (ICCV)}, Oct. 2017, pp. 2961--2969.

\bibitem{mots_trackrcnn}
P.~Voigtlaender, M.~Krause, A.~O\u{s}ep, J.~Luiten, B.~B.~G. Sekar, A.~Geiger,
  and B.~Leibe, ``{MOTS}: Multi-object tracking and segmentation,'' in
  \emph{Proc. IEEE Conf. Comput. Vis. Pattern Recognit. (CVPR)}, Jun. 2019, pp.
  7942--7951.

\bibitem{MOTSFusionJ}
J.~Luiten, T.~Fischer, and B.~Leibe, ``Track to reconstruct and reconstruct to
  track,'' \emph{{IEEE} Robot. Autom. Lett.}, vol.~5, no.~2, pp. 1803--1810,
  Apr. 2020.

\bibitem{camot}
A.~Osep, W.~Mehner, P.~Voigtlaender, and B.~Leibe, ``Track, then decide:
  Category-agnostic vision-based multi-object tracking,'' in \emph{Proc. IEEE
  Int. Conf. Robot. Autom. (ICRA)}, May 2018, pp. 3494--3501.

\bibitem{ciwt}
A.~Osep, W.~Mehner, M.~Mathias, and B.~Leibe, ``Combined image- and world-space
  tracking in traffic scenes,'' in \emph{Proc. IEEE Int. Conf. Robot. Autom.
  (ICRA)}, May 2017, pp. 1988--1995.

\bibitem{beyondpixels}
S.~Sharma, J.~A. Ansari, J.~{Krishna Murthy}, and K.~{Madhava Krishna},
  ``Beyond {P}ixels: Leveraging geometry and shape cues for online multi-object
  tracking,'' in \emph{Proc. IEEE Int. Conf. Robot. Autom. (ICRA)}, May 2018,
  pp. 3508--3515.

\bibitem{motsnet}
L.~Porzi, M.~Hofinger, I.~Ruiz, J.~Serrat, S.~R. Bulo, and P.~Kontschieder,
  ``Learning multi-object tracking and segmentation from automatic
  annotations,'' in \emph{Proc. IEEE Conf. Comput. Vis. Pattern Recognit.
  (CVPR)}, Jun. 2020, pp. 6845--6854.

\bibitem{PointTrack}
Z.~Xu, W.~Zhang, Z.~Tan, W.~Yang, H.~Huang, S.~Wen, and E.~D. nad L.~Huang,
  ``Segment as points for efficient online multi-object tracking and
  segmentation,'' in \emph{Proc. Eur. Conf. Comput. Vis. (ECCV)}, Aug. 2020,
  pp. 264--281.

\bibitem{ReMOTS}
F.~Yang, X.~Chang, C.~Dang, Z.~Zheng, S.~Sakti, S.~Nakamura, and T.~Wu,
  ``{ReMOTS}: Self-supervised refining multi-object tracking and
  segmentation,'' 2020, [{O}nline]. Available: ar{X}iv:2007.03200.

\bibitem{Mahler}
R.~P.~S. Mahler, ``Multitarget {B}ayes filtering via first-order multitarget
  moments,'' \emph{IEEE Trans. Aerosp. Electron. Syst.}, vol.~39, no.~4, pp.
  1152--1178, Oct. 2003.

\bibitem{Vo_gmphd}
B.-N. Vo and W.-K. Ma, ``The {G}aussian mixture probability hypothesis density
  filter,'' \emph{{IEEE} Trans. Signal Process.}, vol.~54, no.~11, pp.
  4091--4104, Oct. 2006.

\bibitem{smcphd}
B.-N. Vo, S.~Singh, and A.~Doucet, ``Sequential {M}onte {C}arlo implementation
  of the {PHD} filter for multi-target tracking,'' in \emph{Proc. Int. Conf.
  Information Fusion (ICIF)}, Jul. 2003, pp. 792--799.

\bibitem{gmphdogm}
Y.~Song, K.~Yoon, Y.-C. Yoon, K.~C. Yow, and M.~Jeon, ``Online multi-object
  tracking with {GMPHD} filter and occlusion group management,'' \emph{IEEE
  Access}, vol.~7, pp. 165\,103--165\,121, Nov. 2019.

\bibitem{eamtt}
R.~Sanchez-Matilla, F.~Poiesi, and A.~Cavallaro, ``Multi-target tracking with
  strong and weak detections,'' in \emph{Proc. Eur. Conf. Comput. Vis.
  Workshops (ECCVW)}, Oct. 2016, pp. 84--99.

\bibitem{gmphdkcf}
T.~Kutschbach, E.~Bochinski, V.~Eiselein, and T.~Sikora, ``Sequential sensor
  fusion combining probability hypothesis density and kernelized correlation
  filters for multi-object tracking in video data,'' in \emph{Proc. IEEE Int.
  Workshop Traffic Street Surveill. Safety Secur. (AVSS)}, Sep. 2017, pp. 1--6.

\bibitem{fu1}
Z.~Fu, P.~Feng, F.~Angelini, J.~Chambers, and S.~M. Naqvi, ``Particle {PHD}
  filter based multiple human tracking using online group-structured dictionary
  learning,'' \emph{IEEE Access}, vol.~6, pp. 14\,764--14\,778, Mar. 2018.

\bibitem{phdgm}
R.~Sanchez-Matilla and A.~Cavallaro, ``A predictor of moving objects for
  first-person vision,'' in \emph{Proc. IEEE Int. Conf. Image Processing
  (ICIP)}, Sep. 2019, pp. 2189--2193.

\bibitem{kcf}
J.~F. Henriques, R.~Caseiro, P.~Martins, and J.~Batista, ``High-speed tracking
  with kernelized correlation filters,'' \emph{{IEEE} Trans. Pattern Anal.
  Mach. Intell.}, vol.~37, no.~3, pp. 583--596, Mar. 2015.

\bibitem{gmphd2012}
V.~Eiselein, D.~Arp, M.~P\"atzold, and T.~Sikora, ``Real-time multi-human
  tracking using a probability hypothesis density filter and multiple
  detectors,'' in \emph{Proc. IEEE Int. Conf. Adv. Video Signal Based Surveill.
  (AVSS)}, Sep. 2012, pp. 325--330.

\bibitem{mtdf}
Z.~Fu, F.~Angelini, J.~Chambers, and S.~M. Naqvi, ``Multi-level cooperative
  fusion of {GM}-{PHD} filters for online multiple human tracking,'' \emph{IEEE
  Trans. Multimedia}, vol.~21, pp. 2277--2291, Sep. 2019.

\bibitem{gmphdn1tr}
N.~L. Baisa and A.~Wallace, ``Development of a {N}-type {GM}-{PHD} filter for
  multiple target, multiple type visual tracking,'' \emph{Journal of Visual
  Communication and Image Representation}, vol.~59, pp. 257--271, 2019.

\bibitem{blob}
R.~H. Evangelio, T.~Senst, , and T.~Sikora, ``Detection of static objects for
  the task of video surveillance,'' in \emph{Proc. IEEE Winter Conf. Appl.
  Comput. Vis. (WACV)}, Jan. 2011, pp. 534--540.

\bibitem{head}
M.~Patzold, R.~H. Evangelio, and T.~Sikora, ``Counting people in crowded
  environments by fusion of shape and motion information,'' in \emph{Proc. IEEE
  Int. Conf. Adv. Video Signal Based Surveill. (PETS Workshop)}, Aug. 2010, pp.
  157--164.

\bibitem{dpm}
P.~F. Felzenszwalb, R.~B. Girshick, D.~McAllester, and D.~Ramanan, ``Object
  detection with discriminatively trained part-based models,'' \emph{IEEE
  Trans. Pattern Anal. Mach. Intell.}, vol.~32, no.~9, pp. 1627--1645, Sep.
  2010.

\bibitem{pose}
E.~Insafutdinov, L.~Pishchulin, B.~Andres, M.~Andriluka, , and B.~Schiele,
  ``Deeper{C}ut: A deeper, stronger, and faster multi-person pose estimation
  model,'' in \emph{Proc. Eur. Conf. Comput. Vis. (ECCV)}, Oct. 2016, pp.
  34--55.

\bibitem{fu2}
Z.~Fu, F.~Angelini, J.~Chambers, and S.~M. Naqvi, ``Multi-level cooperative
  fusion of {GM-PHD} filters for online multiple human tracking,'' \emph{IEEE
  Trans. Multimedia}, vol.~21, no.~9, pp. 2277--2291, 2019.

\bibitem{gmphdntype}
N.~L. Baisa and A.~Wallace, ``Development of a {N}-type {GM-PHD} filter for
  multiple target, multiple type visual tracking,'' \emph{Journal of Visual
  Communication and Image Representation}, vol.~59, pp. 257--271, 2019.

\bibitem{ycyoon1}
Y.-C. Yoon, D.~Y. Kim, Y.~Song, K.~Yoon, and M.~Jeon, ``Online multiple
  pedestrians tracking using deep temporal appearance matching association,''
  \emph{Information Sciences}, vol. 561, pp. 326--351, Jun. 2021.

\bibitem{ycyoon2}
Y.~Song, K.~Yoon, Y.~Yoon, K.~C. Yow, and M.~Jeon, ``Online multi-object
  tracking with historical appearance matching and scene adaptive detection
  filtering,'' in \emph{Proc. IEEE Int. Conf. Adv. Video Signal Based Surveill.
  (AVSS)}, Nov. 2018, pp. 1--19.

\bibitem{kjyoon1}
K.~Yoon, Y.~Song, and M.~Jeon, ``Multiple hypothesis tracking algorithm for
  multi-target multi-camera tracking with disjoint views,'' \emph{IET Image
  Processing}, vol.~12, no.~7, pp. 1175--1184, Jul. 2018.

\bibitem{kjyoon2}
K.~Yoon, D.~Y. Kim, Y.-C. Yoon, and M.~Jeon, ``Data association for
  multi-object tracking via deep neural networks,'' \emph{Sensors}, vol.~19,
  pp. 1--15, Jan. 2019.

\bibitem{cbmot1}
J.~Shen, Z.~Liang, J.~Liu, H.~Sun, L.~Shao, and D.~Tao, ``Multiobject tracking
  by submodular optimization,'' \emph{{IEEE} Trans. Cybern.}, vol.~49, no.~6,
  pp. 1990--2001, Jun. 2019.

\bibitem{cbmot2}
X.~Cao, X.~Jiang, X.~Li, and P.~Yan, ``Correlation-based tracking of multiple
  targets with hierarchical layered structure,'' \emph{{IEEE} Trans. Cybern.},
  vol.~48, no.~1, pp. 90--102, Jan. 2018.

\bibitem{sdp}
F.~Yang, W.~Choi, and Y.~Lin, ``Exploit all the layers: Fast and accurate {CNN}
  object detector with scale dependent pooling and cascaded rejection
  classifiers,'' in \emph{Proc. IEEE Conf. Comput. Vis. Pattern Recognit.
  (CVPR)}, Jun. 2016, pp. 2129--2137.

\bibitem{regionlets}
X.~Wang, M.~Yang, S.~Zhu, and Y.~Lin, ``Regionlets for generic object
  detection,'' \emph{IEEE Trans. Pattern Anal. Mach. Intell.}, vol.~37, no.~10,
  pp. 2071--2084, Oct. 2015.

\bibitem{imagenet}
O.~Russakovsky, J.~Deng, H.~Su, J.~Krause, S.~Satheesh, Z.~H. S.~Ma,
  A.~Karpathy, A.~Khosla, M.~Bernstein, A.~C. Berg, and L.~Fei-Fei,
  ``{I}mage{N}et large scale visual recognition challenge,'' \emph{Int. J.
  Comput. Vis.}, vol. 115, no.~3, pp. 211--252, Dec. 2015.

\bibitem{mapillary}
G.~Neuhold, T.~Ollmann, S.~R. Bul\`o, and P.~Kontschieder, ``The {M}apillary
  {V}istas dataset for semantic understanding of street scenes,'' in
  \emph{Proc. IEEE Int. Conf. Comput. Vis. (ICCV)}, Oct. 2017, pp. 4990--4999.

\bibitem{rrc}
J.~Ren, X.~Chen, J.~Liu, W.~Sun, J.~Pang, Q.~Yan, Y.~Tai, and L.~Xu, ``Accurate
  single stage detector using recurrent rolling convolution,'' in \emph{Proc.
  IEEE Conf. Comput. Vis. Pattern Recognit. (CVPR)}, Jul. 2017, pp. 5420--5428.

\bibitem{bb2segnet}
J.~Luiten, P.~Voigtlaender, and B.~Leibe, ``{PR}e{MVOS}: Proposal-generation,
  refinement and merging for video object segmentation,'' in \emph{Proc. Asian
  Conf. Comput. Vis. (ACCV)}, Dec. 2018, pp. 565--580.

\bibitem{PointNet}
C.~R. Qi, H.~Su, K.~Mo, and L.~J. Guibas, ``Point{N}et: deep learning on point
  sets for 3{D} classification and segmentation,'' in \emph{Proc. IEEE Conf.
  Comput. Vis. Pattern Recognit. (CVPR)}, 2017, p. 652–660.

\bibitem{vot15}
M.~Kristan, J.~Matas, A.~Leonardis, M.~Felsberg, L.~Cehovin, G.~Fernandez,
  T.~Vojir, G.~Hager, G.~Nebehay, and R.~Pflugfelder, ``The visual object
  tracking vot2015 challenge results,'' in \emph{Proc. IEEE Int. Conf. Comput.
  Vis. Workshops (ICCVW)}, Dec. 2015, pp. 1--23.

\bibitem{siamrpn}
B.~Li, J.~Yan, W.~Wu, Z.~Zhu, and X.~Hu, ``High performance visual tracking
  with siamese region proposal network,'' in \emph{Proc. IEEE Conf. Comput.
  Vis. Pattern Recognit. (CVPR)}, Jun. 2018, pp. 8971--8980.

\bibitem{reidsurvey}
M.~Ye, J.~Shen, G.~Lin, T.~Xiang, L.~Shao, and S.~C. Hoi, ``Deep learning for
  person re-identification: A survey and outlook,'' \emph{{IEEE} Trans. Pattern
  Anal. Mach. Intell.}, 2021.

\bibitem{hungarian}
R.~Jonker and A.~Volgenant, ``A shortest augmenting path algorithm for dense
  and sparse linear assignment problems,'' \emph{Computing}, vol.~38, no.~4,
  pp. 325--340, 1987.

\bibitem{ids}
Y.~Li, C.~Huang, and R.~Nevatia, ``Learning to associate: Hybridboosted
  multi-target tracker for crowded scene,'' in \emph{Proc. IEEE Conf. Comput.
  Vis. Pattern Recognit. (CVPR)}, Aug. 2009, pp. 2953--2960.

\bibitem{detectron2}
Y.~Wu, A.~Kirillov, F.~Massa, W.-Y. Lo, and R.~Girshick, ``Detectron2,''
  \url{https://github.com/facebookresearch/detectron2}, 2019.

\bibitem{clearmot}
K.~Bernardin and R.~Stiefelhagen, ``Evaluating multiple object tracking
  performance: The {CLEAR} {MOT} metrics,'' \emph{EURASIP J. Image Video
  Process.}, vol. 2008, pp. 1--10, May 2008.

\bibitem{mhtdam}
C.~Kim, F.~Li, A.~Ciptadi, and J.~M. Rehg, ``Multiple hypothesis tracking
  revisited,'' in \emph{Proc. IEEE Int. Conf. Comput. Vis. (ICCV)}, Dec. 2015,
  pp. 4696--4704.

\bibitem{fwt}
R.~Henschel, L.~Leal-Taixe, D.~Cremers, and B.~Rosenhahn, ``Fusion of head and
  full-body detectors for multi-object tracking,'' in \emph{Proc. IEEE Conf.
  Comput. Vis. Pattern Recognit. Workshops (CVPRW)}, Jun. 2018, pp. 1541--1550.

\bibitem{motdt}
L.~Chen, H.~Ai, Z.~Zhuang, and C.~Shang, ``Real-time multiple people tracking
  with deeply learned candidate selection and person re-identification,'' in
  \emph{IEEE Int. Conf. Multimedia Expo (ICME)}, Jul. 2018, pp. 1--6.

\bibitem{jcc}
M.~Keuper, S.~Tang, B.~Andres, T.~Brox, and B.~Schiele, ``Motion segmentation
  {\&} multiple object tracking by correlation co-clustering,'' \emph{{IEEE}
  Trans. Pattern Anal. Mach. Intell.}, vol.~42, no.~1, pp. 140--153, Jan. 2020.

\end{thebibliography}



\end{document}